\documentclass{article}

\usepackage{titletoc}
\usepackage[final]{corl_2019} 

\usepackage{times}
\usepackage{epsfig}
\usepackage{graphicx}
\usepackage{amsmath}
\usepackage{amssymb}
\usepackage{multirow}
\usepackage{float}
\usepackage{bbm}
\usepackage{enumitem}
\usepackage{pgf}
\usepackage{xifthen}
\usepackage{wrapfig}
\usepackage{capt-of}
\usepackage{mathtools}

\usepackage{xcolor}

\newcommand{\defeq}{\raisebox{-0.15\totalheight}{$\triangleq$}}

\definecolor{linkpink}{rgb}{0.84705882, 0.05882353, 0.49411765}

\title{Learning to Navigate Using Mid-Level Visual Priors}

\author{
  Alexander Sax$^{1,3}$ \hspace{6mm} Jeffrey O. Zhang$^1$ \hspace{6mm} Bradley Emi$^2$ \hspace{6mm} Amir Zamir$^{1,2}$ \vspace{0.6mm} \\ 
  \textbf{Silvio Savarese$^2$ \hspace{8mm} Leonidas Guibas$^{2,3}$ \hspace{8mm} Jitendra Malik$^{1,3}$}
  \vspace{1.5mm} \\
  \normalsize{$^1$University of California, Berkeley \hspace{1mm} $^2$Stanford University \hspace{1mm} $^3$Facebook AI Research} 
  \vspace{1.5mm} \\
  \normalsize{\href{http://perceptual.actor/}{\texttt{\textcolor{linkpink}{http://perceptual.actor}}}}\vspace{-3mm}
}

\begin{document}
\maketitle


\begin{abstract}
How much does having \emph{visual priors} about the world (e.g. the fact that the world is 3D) assist in learning to perform \emph{downstream motor tasks} (e.g. navigating a complex environment)? What are the consequences of not utilizing such visual priors in learning? We study these questions by integrating a generic perceptual skill set (a distance estimator, an edge detector, etc.) within a reinforcement learning framework (see Fig.~\ref{fig:fig1}). This skill set (``mid-level vision") provides the policy with a more processed state of the world compared to raw images.


Our large-scale study demonstrates that using mid-level vision results in policies that \emph{learn faster}, \emph{generalize better}, and achieve \emph{higher final performance}, when compared to learning from scratch and/or using state-of-the-art visual and non-visual representation learning methods. We show that conventional computer vision objectives are particularly effective in this regard and can be conveniently integrated into reinforcement learning frameworks. Finally, we found that no single visual representation was universally useful for all downstream tasks, hence we computationally derive a task-agnostic \emph{set} of representations optimized to support arbitrary downstream tasks. 
 
\end{abstract}




\section{Introduction}
\vspace{-2mm}

The resurgence of deep reinforcement learning (RL) began with a number of nominal works, e.g. the Atari DQN paper~\cite{mnih-dqn-2015} or \emph{pixel-to-torque}~\citep{LevineFDA15}, which collectively showed that RL could be used to train policies directly on raw images. 
Although deep-RL-from-pixels can learn arbitrary policies in an elegant and end-to-end fashion, there are two phenomena endemic to this paradigm: 
\textbf{I.} learning requires massive amounts of data (large sample complexity), and
\textbf{II.} the resulting policies do not transfer well across environments with even modest visual differences (generalization difficulties).

\begin{wrapfigure}{r}{0.47\textwidth}
  \centering
  \definecolor{beige}{rgb}{0.68627451, 0.6627451 , 0.63137255}
      \hspace{-1.0mm}\includegraphics[width=0.48\textwidth]{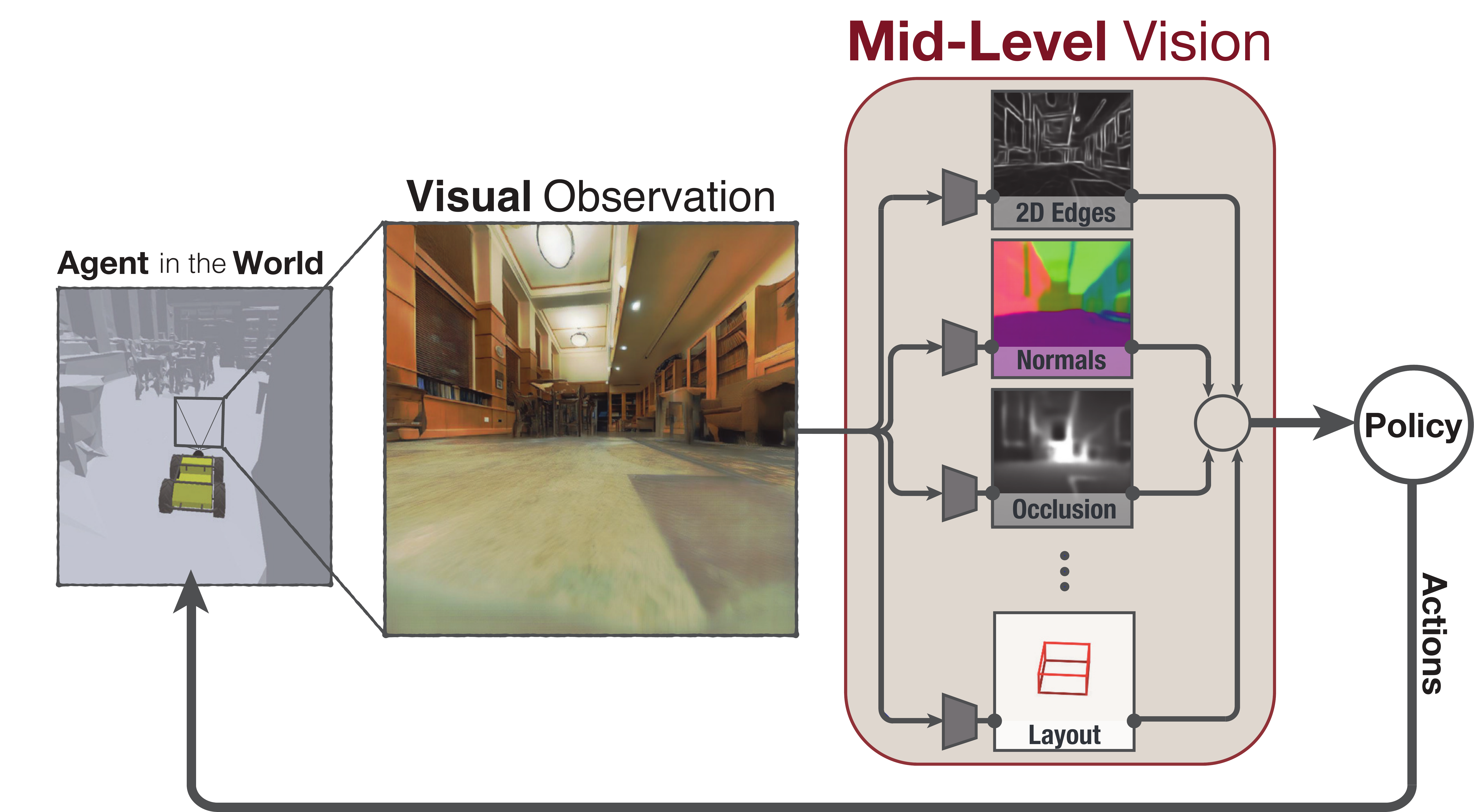}
    \caption{\footnotesize{\textbf{Mid-level vision in an end-to-end framework for learning active robotic tasks.} We report significant advantages in \textit{final performance}, \textit{generalization}, and \textit{sample efficiency} when using mid-level vision, especially compared to learning directly from raw pixels (i.e. bypassing the beige box).}}
    \label{fig:fig1}
  \vspace{-5mm}
\end{wrapfigure}

These two phenomena are characteristic of a type of learning that is \emph{too general}|in that it does not make use of available \emph{priors}. 
In the context of visual perception (the focus of this paper), an example of such priors is that the world is spatially 3D; or there exist certain useful groupings, i.e. ``objects''. These priors are basically \emph{facts} of the world and incorporating them in learning is notably advantageous~\cite{gemanBiasVariance,baxter2000model}. 
That is because the assumption-free style of learning may recover the proper policy but at the expense of massive amounts of data to rediscover such facts (issue \textbf{I}); or may resort to shortcuts spawned by spurious biases in the data to superficially learn faster, which leads to generalization difficulties (issue \textbf{II})~\cite{AmodeiOSCSM16}. 

One of the goals of computer vision is formulating useful visual priors about the world and developing methods for extracting them.
Conventionally, this is done by defining a set of problems (e.g. object detection, depth estimation, etc.) and solving them independently of any ultimate downstream active task (e.g. navigation, manipulation)~\cite{Codevilla2018offline, Anderson2018evalution}. In this paper, we study how such standard vision objectives can be used within RL frameworks as mid-level visual representations~\cite{Peirce2015midlevel}, in order to train effective visuomotor policies.

We show that incorporating mid-level vision can alleviate the aforementioned issues \textbf{I} \& \textbf{II}, resulting in improved \emph{final performance}, \emph{generalization}, and \emph{sample efficiency}. We demonstrate that mid-level vision performs significantly better than SotA state representation learning methods and learning from raw images -- which we find to perform no better than a blind agent when tested on \emph{unseen test data}. Finally, we observe that policies trained using mid-level vision exhibit desirable properties for which they were not explicitly trained, without having to do reward shaping. 


Our study is done using 24 different mid-level visual representations to perform various navigation based downstream tasks in 3 different environments (Gibson~\cite{gibson}, Habitat~\cite{habitat19arxiv}, \emph{ViZDoom}~\cite{vizdoom}). 
Our mid-level vision comes from neural networks trained by existing vision techniques~\cite{taskonomy2018,MITplaces,imagenet} using real images. We use their internal representations as the observation provided to the RL policy. We do not use synthetic data to train the visual estimators nor do we assume they are perfect. 

An interactive tool for comparing various trained policies accompanied with \href{http://perceptual.actor/policy_explorer/}{videos} and \href{http://perceptual.actor/generalization_curves/}{reward curves}, as well as the \href{https://github.com/alexsax/midlevel-reps}{trained models} and \href{#}{code} is available on our \href{http://perception.actor}{website}.

\section{Related Work}
\label{sec:citations}
\vspace{-1mm}

This study has connections to a range of topics, including transfer learning, un/self supervised learning, lifelong learning, reinforcement and imitation learning, control theory, active vision and several others. We overview the most relevant ones within constraints of space.

\textbf{Computer Vision} encompasses approaches that are conventionally designed to solve various stand-alone vision objectives, e.g. depth estimation~\cite{EigenPF14}, object classification~\cite{alexnet}, detection~\cite{Girshick15}, segmentation~\cite{Silberman2012}, pose estimation~\cite{iss, Cao2018openpose}, etc. The approaches use various levels of supervision~\cite{alexnet, NorooziF16, doersch2015unsupervised, bengio2013representation}, but the characteristic shared across these methods is that they are \emph{offline}, i.e. trained and tested on prerecorded datasets and evaluated as a fixed pattern recognition problem. In contrast, the perception of an active agent is fundamentally used in an \emph{online} manner, i.e. the perceptual skill is in service to a downstream goal and the current perceptual decision impacts what the next perceptual observation will follow. Here we study how conventional computer vision objectives can be plugged into such frameworks used for solving downstream active tasks, e.g. navigation.

\textbf{Representation/Feature Learning} shares a goal with our study: to understand how to encode images in a way that provides benefits over using just raw pixels. Many of the most popular representation learning techniques like Variational Autoencoders (VAE)~\cite{kingma2013auto} or alternatives~\cite{Hinton504, Vincent:2008:ECR:1390156.1390294, Matthey2017betaVAELB} are based on Minimum Description Length (MDL) which roughly suggests that ``the best representation is the one that leads to the best compression of the data.'' We show this assumption is not valid, as the most compressive representations are not found to support downstream tasks well.

Other techniques model the dynamics of the environment (e.g. by predicting the next state ~\cite{Jordan1992ForwardMS, vandenOord2018predictive} or by other related objectives~\cite{Dosovitskiy16predicting, curiosity, zhu2007, Raffin2018srltoolbox, AgrawalNAML16}). One advantage of modeling the dynamics is that the representations could useful for planning. These dynamics are not necessarily visual and they may be specialized to the particular morphology or action space of the agent. 

A compendium of several concurrent and recent works have offered supporting evidence that mid-level visual representations could be useful for realistic downstream active tasks: e.g. a specific semantic representation for semantic driving~(\citep{modularityAbstraction, mousavian18, Yang2018ScenePriors}) or dense object descriptors for manipulation~(\citep{denseobjectFlorence2018}).
Independently of our work, \citet{vladlen2019} also studied the role of visual abstractions for learning to act. That work focused on learning policies for synthetic driving and first-person shooter environments; showing that agents equipped with intermediate representations (optical flow, depth, semantic segmentation, and albedo), train faster, achieve higher task performance, and generalize better. While the details of the environments, tasks, and representations differ, the findings are broadly aligned and appear to support each other.

\textbf{Robotics} has long used intermediate visual abstractions such as depth (e.g. SLAM~\cite{RGBDslamreview}), optical flow~\cite{orbslam}, or ground-plane estimation~\cite{dragon2014ground}. However, these usually use the output of the vision solutions in some analytic fashion~\cite{Siciliano2007} which requires the representations to be easy to analyze analytically. When the presesentation is not analytically well understood or in presense of noise (e.g. the latent features from a VAE, noisy depth information) such methods cannot be used. The end-to-end approach has no such constraint, but most end-to-end methods learn with simplistic visual features or \emph{tabula rasa}. This paper studies the utility of incorporating mid-level visual features in a learning-based end-to-end frameworks.



\section{Methodology}
\label{sec:methods}

\begin{figure*}
\vspace{-3mm}
\hspace{-3.0mm} \includegraphics[width=1.05\textwidth]{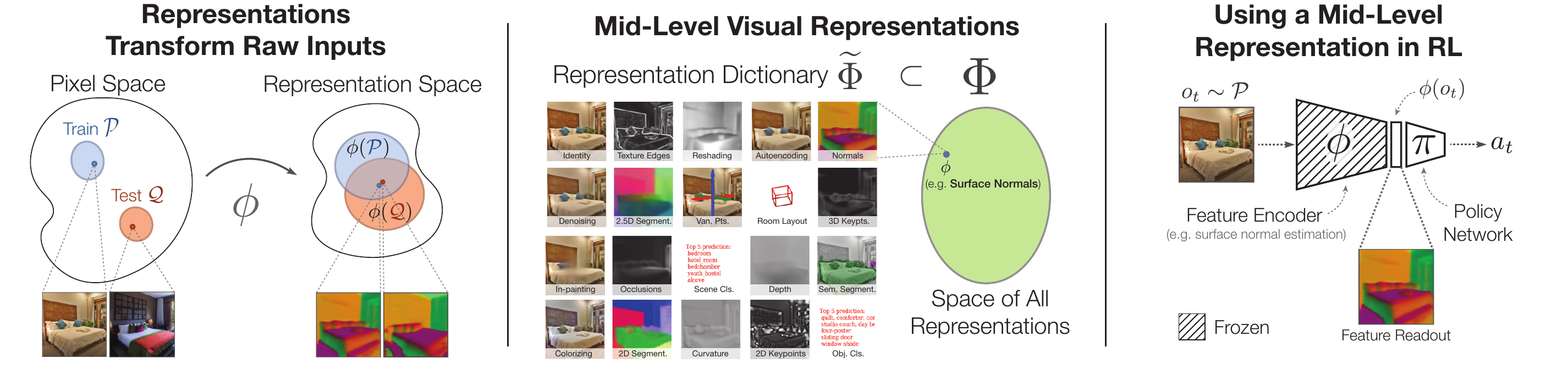}
\vspace{-5mm}
   \caption{\footnotesize{\textbf{Illustration of our approach.} \textit{Left:} The job of a feature is to warp the input distribution, potentially making the train and test distributions look more similar to the agent. \textit{Middle:} Visualizations for 19 of 24 mid-level vision objectives in our dictionary \raisebox{-1pt}{$\widetilde{\Phi}$}. The dictionary is a sample of all possible transforms ($\Phi$), and the best function for a given task must have the proper ``invariance'', i.e. ignore irrelevant parts of the input while retaining the information required for solving the downstream task. \textit{Right:} Representations from fixed encoder networks are used as the observation for training policies in RL.}}
   \label{fig:hypothesis_show}
\vspace{-3mm}
\end{figure*}

Our goal is to study the role of mid-level representations of raw visual data toward performing downstream robotic tasks better. More concretely, the goal is to maximize an agent's performance in a test setting ($\mathcal{Q}$) that is similar \emph{but not identical to the training distribution}, $\mathcal{P}$. Our setup assumes access to a set of functions $\Phi = \{\phi_1, \ldots, \phi_m\}$ that can be used to transform raw sensory data into some (possibly useful) representation. We define $\mathcal{P}_\phi~ \defeq~\phi(\mathcal{P})$ as image of the the training distribution under $\phi$  as and $\mathcal{Q}_\phi~ \defeq~\phi(\mathcal{Q})$ analogously. This shown in Fig.~\ref{fig:hypothesis_show}, left). In this paper we show that when this dictionary is a set of mid-level visual representations, agents can achieve better final performance, generalization, and sample efficiency.


\subsection{Using Mid-Level Vision for Active Tasks: The Role of Representations}
\vspace{-7pt}

Why could a visual feature, e.g. image $\rightarrow$ surfance normals, improve test-time performance on a downstream task, compared to simply using the image raw? A good representation $\phi$ transforms the inputs in a way that preserves the task-relevant information while eliding task-irrelevant differences between the train and test distributions. Roughly speaking, a good representation makes $\mathcal{P}_\phi \approx \mathcal{Q}_\phi$ without affecting the training reward ($R_{\mathcal{P}_\phi}$), as illustrated in Fig.~\ref{fig:hypothesis_show}-left. In this way, using RL to maximize the training reward also improves the test-time performance, $R_{\mathcal{P}_\phi \rightarrow \mathcal{Q}_\phi}$.


Our mid-level representations come from a set of neural networks that were each trained, offline, for a specific vision objective~\cite{taskonomy2018}~(see Fig.~\ref{fig:hypothesis_show}-middle). We freeze each encoder's weights and use the network ($\phi$) to transform each observed image $o_t$ into a summary statistic $\phi(o_t)$ that we feed to the agent. During training, only the agent policy is updated (as shown in Fig.~\ref{fig:hypothesis_show}-right). Freezing the encoder networks has the advantage that we can reuse the same features for new active tasks without degrading the performance of already-learned policies. This approach is almost certainly not ideal, but agents trained using mid-level representations in this way still outperform the current SotA.

\subsection{Core Analysis: Final Performance, Generalization, and Rank Reversal}\label{sec:core_analysis}

\vspace{-7pt}
\textbf{Final Performance:}
We evaluate agents in a \emph{test space} distinct from where the agents were trained. Maintaining a train/test split is crucial, and we found that training performance was not necessarily predictive of test performance.

\vspace{-2pt}
\textbf{Generalization:}
We provide a detailed analysis of how different agents generalize|reporting both the common metric of generalization (the gap between train and test buildings) and alternatives (e.g. performance of test episodes significantly longer or harder than the training episodes). 

\vspace{-2pt}
\textbf{Sample Complexity:}
We examine whether an agent equipped with mid-level vision can learn faster than a comparable agent that learns \textit{tabula rasa} (i.e. with vision, but no priors about the world). We report sample complexity both in terms of the number of training updates/frames (the usual metric), and also as a function of how many buildings (sampling clusters) are in the training set.

\subsection{Rank Reversal: Which Mid-Level Feature to Use?}
\vspace{-7pt}
\textit{Can a \textbf{single} feature support all downstream tasks? Or is a set of features required for gaining these feature benefits in arbitrary tasks?}
We demonstrate that no feature is universally useful for all downstream tasks (Sec.~\ref{sec:rank_reversal}). We show this by demonstrating cases of \textit{rank-reversal}|when the ideal features for one task are non-ideal for another task (and vice-versa). 
Formally, for tasks $T_0$ and $T_1$ with best features $\phi_0$ and $\phi_1$ respectively, where the test reward for each task is denoted $R_{i} \defeq R_{\mathcal{P}_i \rightarrow \mathcal{Q}_i}$ we show that
$R_{0}(\phi_0) > R_{0}(\phi_1)~\text{and}~R_{1}(\phi_0) < R_{1}(\phi_1)$.
For instance, we find that \emph{depth estimation} features perform well for visual exploration and \emph{object classification} for target-driven navigation, but neither do well vice-versa.

\subsection{Max-Coverage Feature Set for Arbitrary Tasks}
\vspace{-7pt}
As a consequence of the rank-reversal phenomenon, one needs to select the mid-level feature based on the current downstream task of interest, and keep updating it every time that task changes. In this section we ask: when the downstream task is unknown or changes, could a predetermined set of features provide better worst-case (generic) perception than using any single feature? We give an example of such a set: the \emph{Max-Coverage Feature Set}, which is a minimal covering set of the space of useful visual abstractions. In Sec.~\ref{section:max_coverage}, we demonstrate that this set can capture the benefits of mid-level vision, comparable to if we had known the best feature \emph{a priori}. Finding the Max-Coverage (M-C) feature set can be formulated as a sequence of $O(\log|\text{\raisebox{-0.15\totalheight}{$\widetilde{\Phi}$}}|)$ Boolean Integer Programs and solved efficiently in $<4$ seconds. Since the main goal of our study is to examine the utility of mid-level vision and to demonstrate that no single feature is universal, we provide the detailed formulation and analysis of the Max-Coverage feature set in Appendix \ref{apx:max_coverage_feature_set_formulation}. 

\section{Case Study: Vision-Based Navigation}
We adopt a class of active tasks (navigation) and apply the described methodology to perform our study. 
The study examines representations driven by 24 different mid-level objectives, compared against 8 state-of-the-art baselines and 3 separate controls, all on 3 distinct navigation tasks. The remainder of this section describes our experimental setup, and complete details are in the appendix.



\textbf{Environments:} Our experimental setup is the same as in the CVPR 19 Habitat Challenge~\cite{habitat19iccv}; we use the Habitat platform~\cite{habitat19arxiv} with the Gibson~\cite{gibson} dataset.  The dataset captures the intrinsic visual and semantic complexity of real-world scenes by scanning 572 actual buildings. See our \href{http://perceptual.actor}{website} for videos of the trained policies generalizing to real robots (with no finetuning).

To establish the universality of the results, besides Habitat, we also perform the study using two other environments: \emph{Gibson Environments}~\cite{gibson} (a visually realistic simulator integrated with PyBullet dynamics and operating on Gibson dataset) and \emph{ViZDoom}~\cite{vizdoom} (a 3D first-person game).

\begin{figure}[H]
    \centering
    \vspace{-2mm}
    \hspace{-0mm}\includegraphics[width=0.9\textwidth,trim=0cm 0cm 0cm 1.0cm, clip=true]{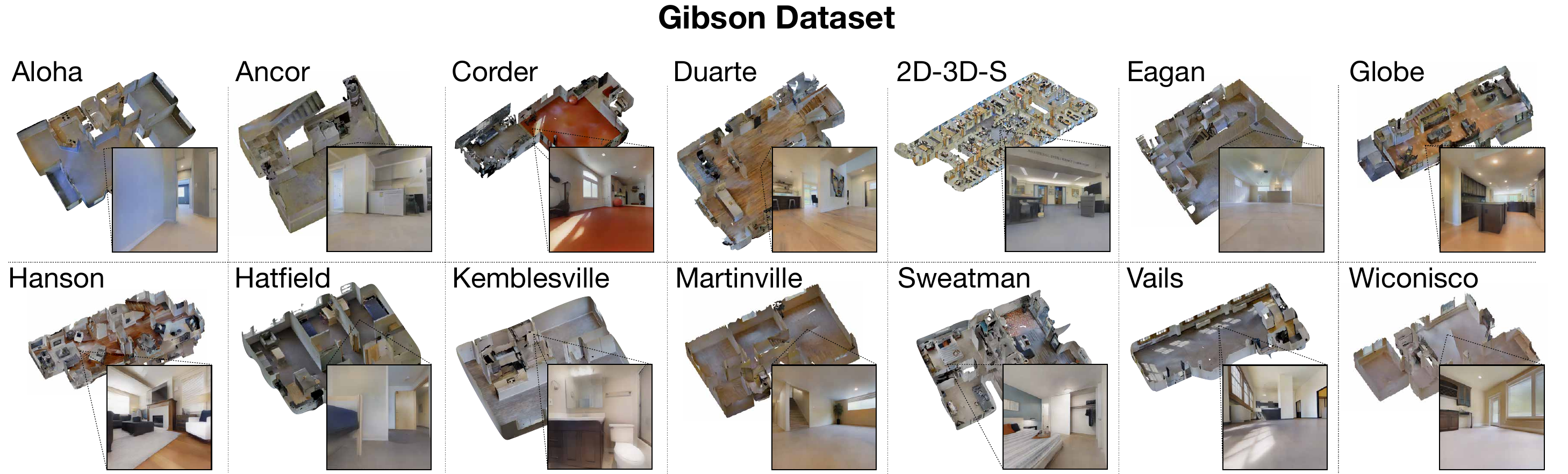}
   \vspace{-2mm}
   \caption{\footnotesize{\textbf{Sample of 14 buildings from the Gibson~\cite{gibson} dataset.} One floor is shown from each space. Boxed images indicate sample observations from the Gibson~\cite{gibson} environment, while observations in Habitat~\cite{habitat19arxiv} come directly from the (shown) mesh textures. We use 72 buildings for training and 14 for testing.}}
   \vspace{-2mm}
   \label{fig:meshes}
\end{figure}


\textbf{Train/Test Split:}
We train and test our agents in two disjoint sets of buildings (Fig.~\ref{fig:meshes}); test buildings are completely unseen during training.  We use up to 72 building for training and 14 test buildings for testing.  The train and test spaces comprise 15678.4$m^2$ (square meters) and 1752.4$m^2$, respectively.

\subsection{Downstream Navigation Tasks}\label{sec:active_tasks}
\vspace{-7pt}
We present our case study using three common and useful navigation-type tasks: \textit{local planning}, \textit{visual exploration}, and  \textit{navigation to a visual target} (see videos  \href{http://perceptual.actor/policy_explorer/}{here}); described below (detailed in Appendix \ref{sec:supmat_tasks}). 

\begin{description}[leftmargin=2mm]
\small
\vspace{-1mm}
\small
\item\textbf{Local Planning (aka Point Goal~\cite{anderson2018evaluation,habitat19arxiv}):}
The agent must direct itself to a given nonvisual target destination (specified using coordinates), avoiding obstacles and walls as it navigates. This skill might be useful for traversing sparse waypoints along a desired path. 
During training, the agent receives a large one-time reward for reaching the goal, and in the dense-reward variant, also receives positive reward proportional to the progress it makes (in Euclidean distance) toward the goal. There is also a small negative reward for living. The maximum episode length is 500 timesteps, and the target location is 1.4 to 15 meters from the start.

\small
\item\textbf{Visual Exploration:}
The agent must visit as many \textbf{new} parts of the space as quickly as possible. The space is partitioned into small occupancy cells that the agent ``unlocks'' by scanning with a myopic laser range scanner. This scanner reveals cells directly in front of the agent for up to 1.5 meters. The reward at each timestep is proportional to the number of newly revealed cells. The episode ends after 1000 timesteps. 

\item\textbf{Navigation to a Visual Target:}
In this scenario the agent must locate a specific target object (a wooden crate) as fast as possible with only \textit{sparse rewards}. Upon touching the target there is a large one-time positive reward and the episode ends. Otherwise there is a small penalty for living. The target (wooden crate) is fixed between episodes although the agent must learn to identify it. The location and orientation of both the agent and target are randomized. The maximum episode length is 400 timesteps, and the shortest path is usually over 30.

\end{description}

\vspace{-3pt}

\textbf{Sensory Input:}
For each task, the sensory observation space contains RGB images of the onboard camera. In addition, the \emph{minimum} amount of side information needed to feasibly solve the downstream task are appended; that is the target direction/location for local planning, unlocked occupancy cells for exploration, and nothing for Visual Target Navigation. Unlike the common practice, we do not include proprioception information such as the agent's joint positions, velocities, or any other unessential side information in order to strictly test the perceptual skills of the agents.

\vspace{-3pt}

\textbf{Action Space:}
We assume a low-level controller for robot actuation, enabling a high-level action space of $\mathcal{A}=\small{\{}\texttt{\small{turn\_left}(10$^\circ$)},~\texttt{\small{turn\_right}(10$^\circ$)},~ \texttt{\small{move\_forward}(0.25$m$)}\small{\}}$.

\subsection{Reinforcement Learning Algorithm}
\vspace{-7pt}


We use an off-policy variant of Proximal Policy Optimization~\cite{PPO} (PPO, details in Appendix~ \ref{apx:ppo}), with a small controller network.
For each task and each environment we conduct hyperparameter searches optimized for \textit{scratch} and all the state-of-the-art baselines (see section~\ref{section:baselines}). For our mid-level agents, we use the same hyperparameters that were optimized for scratch.




\subsection{Mid-Level Representations}
\vspace{-7pt}
For our experiments, we used representations derived from one of 24 different computer vision objectives (Fig.~\ref{fig:hypothesis_show}). This set covers various common modes of computer vision objectives: from texture-based (e.g. denoising), to 3D pixel-level (e.g. depth estimation), to low-dimensional geometry (e.g. room layout), to semantic (e.g. object classification). For these objectives we used the networks of~\cite{taskonomy2018} trained on a dataset of 4 million static images of indoor scenes~\cite{taskonomy2018}. 
All networks were trained with identical hyperparameters and using a ResNet-50~\cite{resnet} encoder.
For a full list of vision objectives and samples of the networks evaluated in our environments, see Appendix \ref{sec:task_dict}. 

\subsection{State-Representation Baselines}\label{section:baselines}
\vspace{-7pt}
We include multiple control groups in order to address possible confounding factors, and we compare against several state-of-the-art baselines. We describe the most important ones here and defer remaining descriptions to Appendix~\ref{apx:baseline_description}. 
\vspace{-1.5mm}
\begin{description}[leftmargin=2mm]
\small{
\item \textbf{Tabula Rasa (Scratch) Learning:}
The most common approach, \emph{tabula rasa} learning trains the agent from \emph{scratch}. In this condition, the agent receives the raw RGB image as input and uses a randomly initialized AtariNet~\cite{mnih-dqn-2015} tower that is trained with the policy.

\vspace{-0.5mm}
\item \textbf{Blind Intelligent Actor:}
The \emph{blind} baseline is the same as \emph{tabula rasa} except that the visual input is a fixed image and does not depend on the state of the environment. The \emph{blind} agent is a particularly informative and crucial baseline since it indicates how much performance can be squeezed out of the nonvisual biases, correlations, and overall structure of the environment. For instance, in a narrow straight corridor which leads the agent to the target, there should be a small performance gap between \emph{sighted} and \emph{blind}. 



\vspace{-0.5mm}
\item \textbf{State-of-the-Art Representation Learning:} We compare against several state-of-the-art representation-learning methods, including \emph{dynamics-modeling} ~\cite{Munk2016Forward, ShelhamerMAD16Inverse, Jordan1992ForwardMS}, \emph{curiosity}~\cite{curiosity}, \emph{DARLA}~\cite{higgins2017darla}, and \emph{ImageNet pretraining}~\cite{alexnet}, enumerated in Figure~\ref{fig:final_perf}.

\item \textbf{Non-Learning:} Methods such as SLAM~\cite{RGBDslamreview} have long used intermediate representations such as depth, but inside a specialized non-learning framework. SLAM may not always be applicable, but when it is it shows how much of the problem can solved with hand-engineered systems.

}
\end{description}

\section{Experimental Results}\label{sec:results}
\vspace{-1mm}

This section presents results from a case study of agents trained with mid-level representations. In the main paper we primarily focus on the \emph{Local Planning} task performed in Habitat~\cite{habitat19arxiv}. The supplementary material provides the results of the experiments in other environments (Gibson, Doom) and with other downstream tasks (exploration, visual target navigation), which show the same trends.

\subsection{Final Performance: Mid-Level Features Exhibit Better Performance}\label{section:performance}
\vspace{-3pt}

\noindent\textbf{Higher reward on the test set:} 
Agents using mid-level visual representations achieve performance significantly higher than agents trained from scratch (Fig.~\ref{fig:final_perf}, left). This was tested in both Gibson~\cite{gibson} environment and Habitat~\cite{habitat19arxiv}, as shown in Fig~\ref{fig:final_perf}. The Spearman's $\rho$ between performance in Habitat and Gibson was 0.87, indicating a high degree of agreement between rankings in the two environments. Notably, \emph{\textbf{scratch} and several of the SotA features do not perform much better than a \textbf{blind} agent in the test space}, which suggests they heavily overfitted to the training set (Appendix~\ref{apx:additional_generalization_scratch_vs_blind}).

\definecolor{midlevel}{rgb}{0.67, 0.05333333, 0.10333333}

\begin{figure}[H]
    \centering
    \vspace{-2mm}
    \hspace{-0mm}\includegraphics[width=0.9\textwidth]{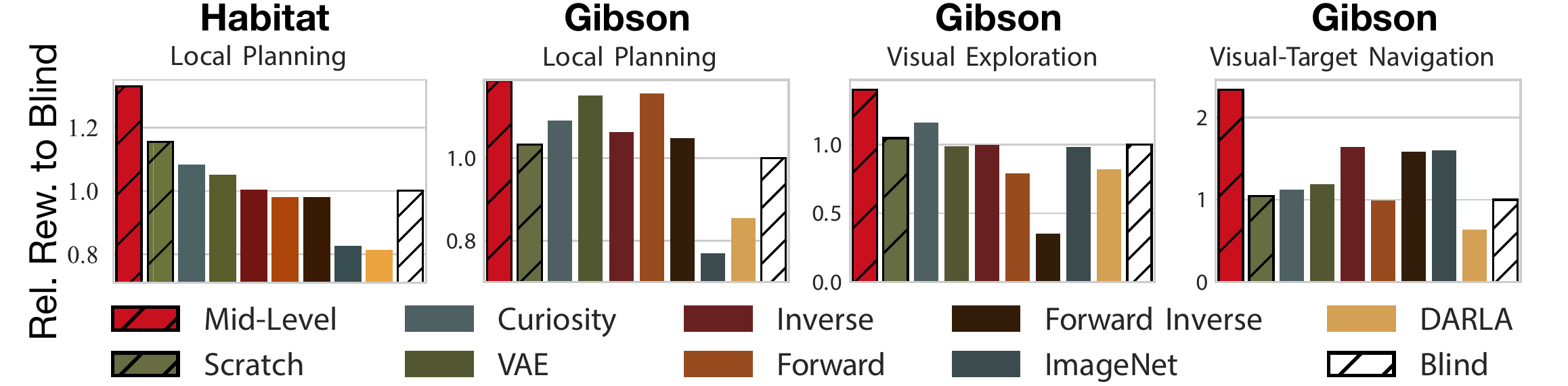}
    \vspace{-2mm}
    \caption{\footnotesize{\textbf{Agents using mid-level vision had higher reward on the test set.} Each bar plot compares average test-set reward on a different task. On every task, agents using \textcolor{midlevel}{mid-level vision} outperformed those learning \emph{from scratch} or using alternative SotA representation-learning methods. Significance tests in Appendix \ref{apx:additional_generalization}.}}
    \label{fig:final_perf}
    \vspace{-3mm}
\end{figure}

\begin{figure}[H]
    \definecolor{seabornblue}{rgb}{0.14509804, 0.46666667, 0.70588235}
    \definecolor{seabornorange}{rgb}{0.98039216, 0.49803922, 0.05490196}
    \definecolor{seaborngreen}{rgb}{0.17254902, 0.62745098, 0.16862745}
    \definecolor{seabornpurple}{rgb}{0.56078431, 0.42352941, 0.74117647}
    \vspace{-2mm}
    \hspace{-1mm}\includegraphics[width=1.02\columnwidth]{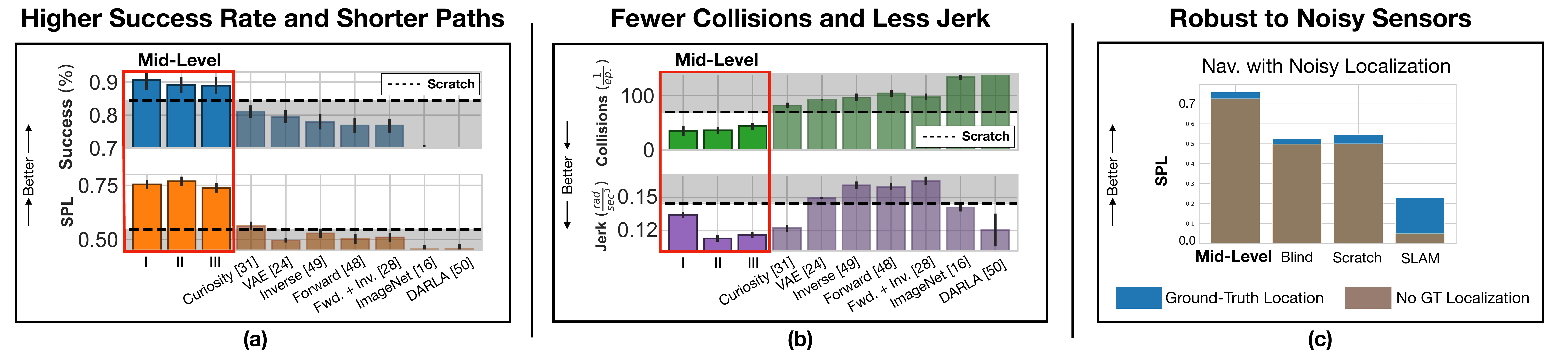}
    \vspace{-7mm}
    \caption{\footnotesize{\textbf{Desirable emergent behavior and robustness under uncertainty.} \textit{(a),(b):} The colorful bar charts show that agents using mid-level vision learn desirable behavior that is not explicitly coded for in the reward: they experience fewer \textcolor{seabornpurple}{collisions}, less \textcolor{seaborngreen}{jerk}
    \protect\footnote{Acceleration is similar and shown in Appendix~\ref{apx:additional_analysis}}, and achieve a higher \textcolor{seabornblue}{success rate} and shorter average \textcolor{seabornorange}{path length}.
    \textit{(c):} Removing ground-truth agent localization (gap between the blue and brown rectangles) harms localization and hurts classical methods and \emph{scratch} more than mid-level agents.
    }}  
    \vspace{-4mm}
    \label{fig:final_perf_other_metrics}
\end{figure}

\noindent\textbf{Desirable emergent behavior without reward engineering:} Although agents were trained only to maximize training reward, we found that mid-level-vision-based agents exhibited other desirable properties such as \emph{fewer collisions, less acceleration and jerk, and better performance on alternative task metrics}, as shown in Fig. \ref{fig:final_perf_other_metrics} (a,b)\footnote{The top mid-level features (I, II, and III) were chosen based on reward (see Sec.~\ref{sec:rank_reversal}). Error bars = 1 SE.}. Other papers~\cite{bansal2019} have specifically built in this desirable behavior, but we find that simply adding in appropriate perception goes a long way towards fixing the issue without having to hand engineer the reward. 

\vspace{-3pt}
When using mid-level vision, agents performed equally well with sparse or dense rewards (0.74 vs 0.76 SPL)\footnote{SPL = Success Weighted by Path Length (as in~\cite{anderson2018evaluation}, see Appendix~\ref{sec:supmat_metrics})} and significantly outperformed SotA methods--\emph{even when the SotA methods used dense reward and/or were explicitly engineered to handle sparse reward} (e.g.~\citet{curiosity}: 0.56 SPL).

\noindent\textbf{More robust agents:}  Fig.~\ref{fig:final_perf_other_metrics} (c) demonstrates that mid-level vision-based agents robustly handle noisy inputs. We removed agents' access to ground-truth agent localization (via an inertial measurement unit) and found that \textbf{(1)} the gains from using mid-level priors were much larger than the gains from using a (ground-truth) map (+0.23 SPL and +0.05 SPL vs. \emph{scratch}), \textbf{(2)} mid-level agents without GT localization still outperformed other approaches \emph{even when those used a map} (0.73 vs. 0.55 SPL). \textbf{(3)} classical approaches such as SLAM exhibited poor performance with estimated (instead of ground-truth) localization (0.05 SPL) or depth (0.23 SPL).


\subsection{Generalization: Mid-Level Features Generalize Better to New Domains and Distant Goals}\label{section:generalization}
\vspace{-1mm}

In the previous section we examined test-set reward, which is a determined by how well the agent learns on the training set how well that generalizes to the test set. We find that agents using mid-level features generalize well: both when the training and test buildings are drawn from the same distribution, and also when the test episodes are much longer than anything seen during training.


\begin{wrapfigure}{r}{0.5\textwidth}
    \centering
    \vspace{-1.5mm}
    \hspace{-0.5mm} \includegraphics[width=0.49\textwidth]{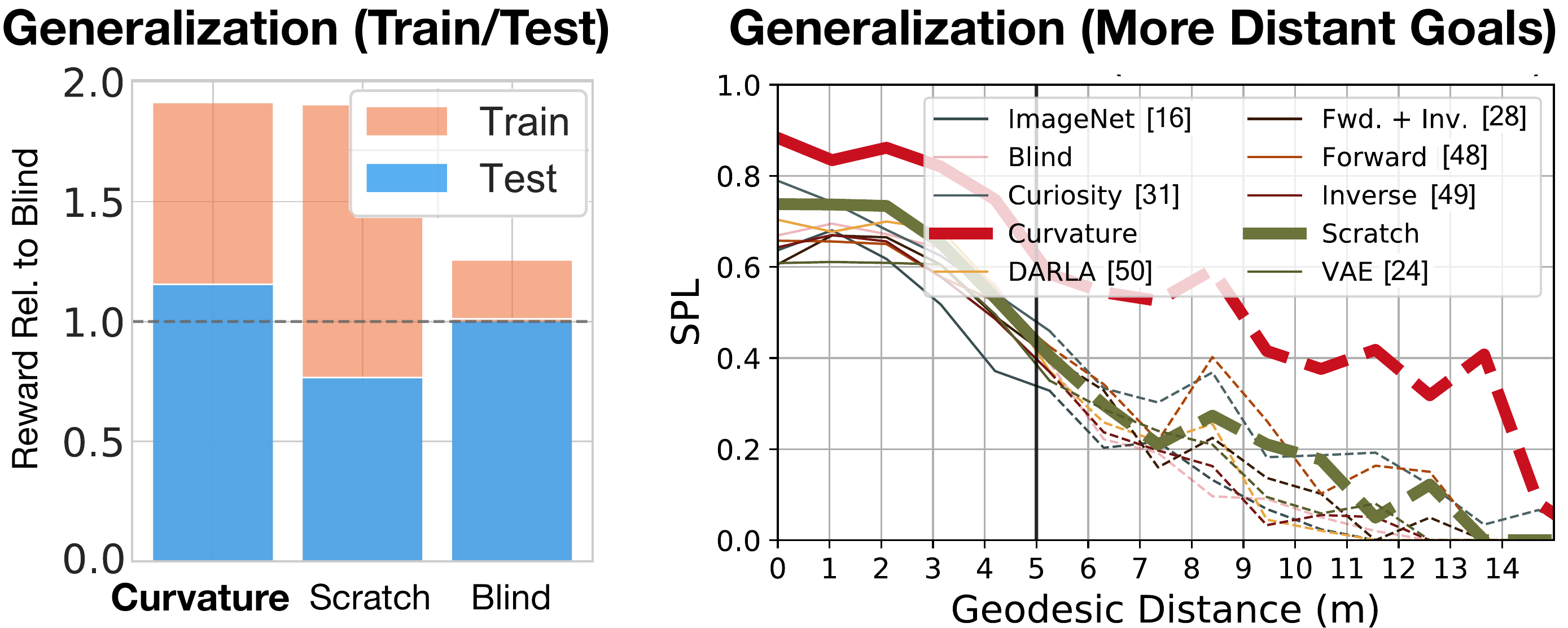}
    \vspace{-2mm}
    \caption{\footnotesize{\textbf{Agents using mid-level vision generalize to new buildings and more distant goals.} \textit{Left:} While both \emph{curvature} and \emph{scratch} achieve the same reward on the training set (4 buildings), agents using mid-level vision generalize far better than \emph{scratch}, which performs worse than a \emph{blind} agent (gray line). \textit{Right:} Agents using \emph{curvature} features retain performance on episodes longer than anything seen during training (right of black line); outperforming alternative approaches.}}
    \vspace{-4mm}
    \label{fig:generalization}
\end{wrapfigure}

\textbf{Train/Test in Different Buildings:} To analyze generalization in more detail, we performed a study with a notably smaller training set (4 training buildings), as a smaller training set provides a larger opportunity for overfitting and makes it more obvious which methods are prone to it. As Fig.~\ref{fig:generalization} (left) shows, both \emph{scratch} and \emph{mid-level} agents achieve similar reward on the training data, but agents using mid-level visual representations generalize better to new buildings (SPL of 0.65 vs. 0.46).


\textbf{Generalizing to Longer Episodes:} Another common definition of generalization rests on whether agents can extrapolate to novel situations unseen in the training set. Fig.~\ref{fig:generalization} (right) shows that agents using mid-level vision are better able to extrapolate to episodes longer than those seen during training (under $5m \rightarrow$ over $5m$), as compared to agents trained from \emph{scratch} (\emph{scratch}: 0.70 SPL $\rightarrow$ 0.33 SPL vs. \emph{curvature}: 0.84 SPL $\rightarrow$ 0.56 SPL). Agents using mid-level vision even significantly outperformed agents that explicitly model the environment dynamics (0.68 $\rightarrow$ 0.37 SPL), whereas a dynamics representation is expected to help with this type of generalization.


\subsection{Sample Complexity: Mid-Level Visual Representations Result in Learning Faster}\label{section:sample_efficiency}

\vspace{-3pt}


We find that agents using mid-level visual representations need less data. In our case, about an order of magnitude less. We report two different quantities to support this claim.

\begin{wrapfigure}{r}{0.45\textwidth}
    \centering
    \vspace{-4.5mm}
    \includegraphics[width=0.45\textwidth]{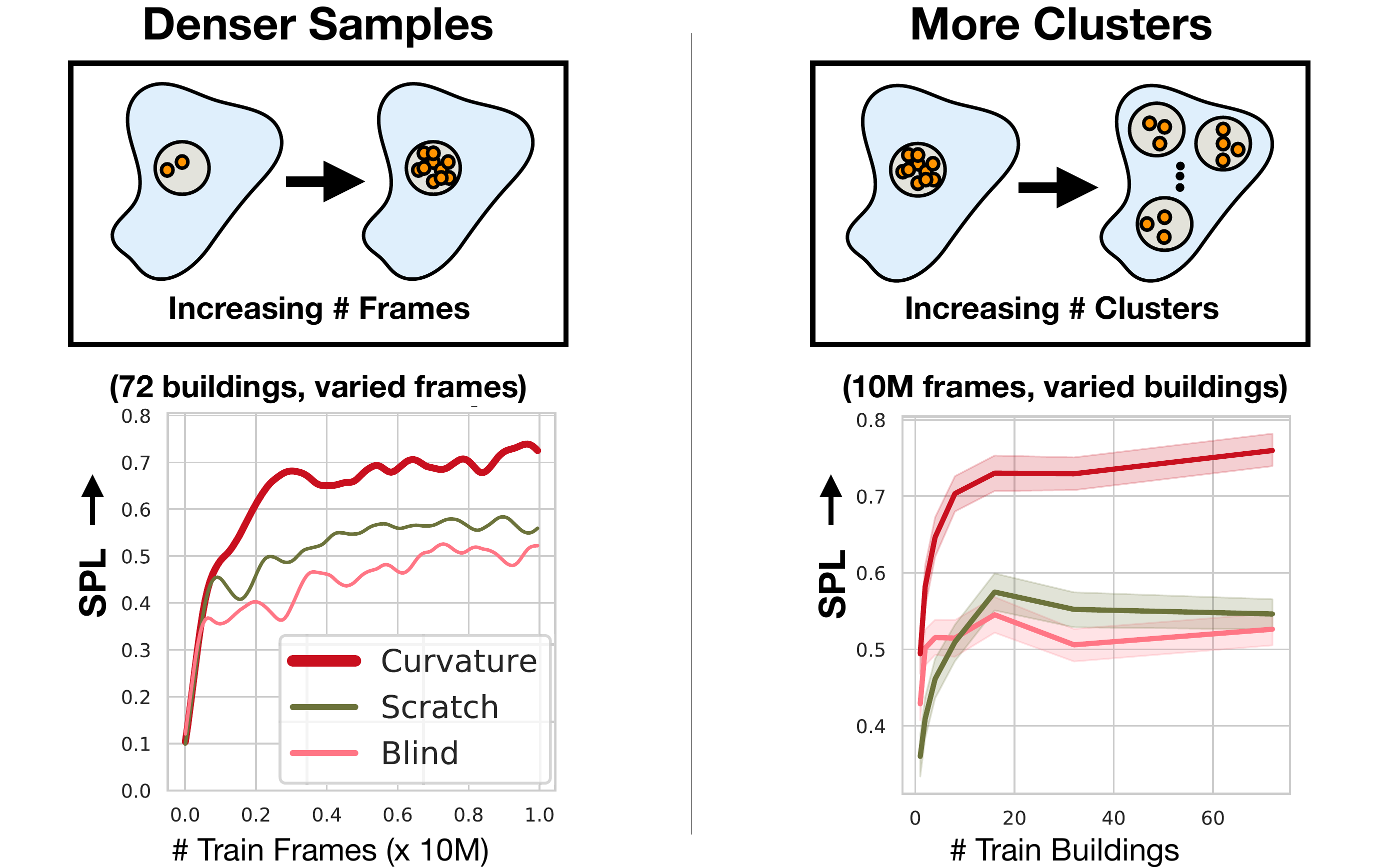}
    \vspace{-5mm}
    \caption{\footnotesize{\textbf{Mid-level vision-based agents learn with fewer frames and fewer buildings.} \textit{Left:} They achieve the final reward of agents trained \emph{tabula rasa} in 15\% of number of frames. Right: Agents with mid-level vision use fewer buildings (11\%) to reach same performance as \emph{scratch}.}}
    \vspace{-5,5mm}
    \label{fig:sample_complexity}
\end{wrapfigure}

\textbf{Performance by Number of Training Frames:} We report the performance vs. number of training frames that the agent receives from the environment. Mid-level-vision-based agents achieve the final maximum reward of agents trained \emph{tabula rasa} in 15\% of the time (Fig.~\ref{fig:sample_complexity}-left).

\textbf{Performance by Number of Sample Clusters (Buildings):}
One rarely noted but critical factor when measuring the number of training frames is that the frames are highly correlated. In our case, this implies two frames coming from the same building supply less diversity/visual learning value compared to two frames coming from two different buildings.
Therefore, we also measure the performance as the number of sample clusters (buildings) increases, as opposed to just the frame count (Fig.~\ref{fig:sample_complexity}, right). With only 11\% (8/72) of the training buildings, agents using mid-level priors achieve the same (or better) reward as \emph{scratch} does when trained on the 100\% dataset .

\subsection{No Universal Feature: Rank Reversal in Navigation and Exploration}\label{sec:rank_reversal}
\vspace{-5pt}

Our results suggest there are not one or two canonical representations that consistently outperform all else. Instead, we find that the choice of representation depends upon the downstream task (Tab.~\ref{fig:rank_reversal_table}, top). For example, the top-performing exploration agent used representations for \emph{Distance Estimation}, perhaps because an effective explorer needs to identify large open spaces. In contrast, the top navigation agent used representations for \emph{Object Classification}---ostensibly because the agent needs to identify the target crate. Despite being top of their class on their preferred tasks, neither representation performed particularly well on the other task. Using 10 seeds per representation per task, this result was statistically significant (in both directions) at the $\alpha=0.0005$ level.

\begin{wrapfigure}{r}{0.47\textwidth}
    \centering
   \begin{minipage}{0.48\textwidth}
        \vspace{-3.5mm}
        \setlength{\tabcolsep}{2pt}
        \centering
        \scriptsize{\textbf{Per-Task Top Feature (Reward)}}
        \scriptsize
        \begin{tabular}{|lll|}
            \hline 
            
            \scriptsize{\textbf{Navigation}} & \scriptsize{\textbf{Exploration}} & \scriptsize{\textbf{Local Planning}} \\
            \hline 
            1. Obj. Class.~(5.90) & 1. Distance~(5.90) & 1. 3D Keypts.~(15.5) \\
            2. Sem. Segm.~(5.86) & 2. Reshading~(5.79) & 2. Normals~(15.1) \\
            3. Curvature~(4.74) & 3. 2.5D Segm.~(5.60) & 3. Curvature~(14.8) \\
            \hline 
    	\end{tabular}
    \end{minipage}
    \begin{minipage}{0.47\textwidth}
        \vspace{2mm}
        \setlength{\tabcolsep}{6pt}
        \scriptsize
        \centering
        \textbf{Correlation (Spearman's $\rho$)}
        \scriptsize
        \begin{tabular}{|l|ccc|}
            \hline 
            
            & \textbf{Nav.} & \textbf{Exp.} & \textbf{Plan.} \\
            \hline 
            \textbf{Nav.} & - & -0.09 & 0.15 \\
            \textbf{Exp.} & -0.09 & - &  0.07  \\
            \textbf{Plan.} & 0.15 &  0.07  & - \\
            \hline 
    	\end{tabular}
        \vspace{-1mm}
        \captionof{table}{\footnotesize{\textbf{Feature ranks are uncorrelated between tasks.}} \textit{Top:} Top 3 features per task (Gibson). Note that no feature is consistently on top. \textit{Bottom:} Cells show Spearman's $\rho$ between feature ranks on different tasks; no correlation was statistically significant.}
        \label{fig:rank_reversal_table}
    \end{minipage}
    \vspace{-2mm}
\end{wrapfigure}

    \definecolor{navigation}{rgb}{0.19215686, 0.56078431, 0.89019608}
    \definecolor{exploration}{rgb}{0.98039216, 0.43529412, 0.38823529}

The above pair is an example of rank reversal, which is actually a common phenomenon. It is so common that the feature rankings were effectively \textbf{uncorrelated} among our three tasks (Table~\ref{fig:rank_reversal_table}, bottom), even though the three tasks seem superficially similar (all locomotion-based). Since the within-task ranks were consistent ($\rho=0.87$ between environments), the choice of task was the primary determining factor for feature rank. Moreover, that choice determined whether whole families of related features would perform well. We found that semantic features were useful for navigation while geometric features were useful for exploration; and the \textcolor{navigation}{semantic}/\textcolor{exploration}{geometric} characterization was extremely predictive of final performance. We quantify the strong statistical significance in both Gibson and Doom using 120 pairwise significance tests 
in Appendix~\ref{apx:supmat_rank_reversal}.




\vspace{-1mm}   
\subsection{When the Downstream Task is Unknown or Changing: Use a Set of Features}\label{section:max_coverage}
\vspace{-5pt}

\begin{wrapfigure}{r}{0.45\textwidth}
        \vspace{-3.5mm}
        \hspace{-2.5mm}\includegraphics[width=0.48\textwidth]{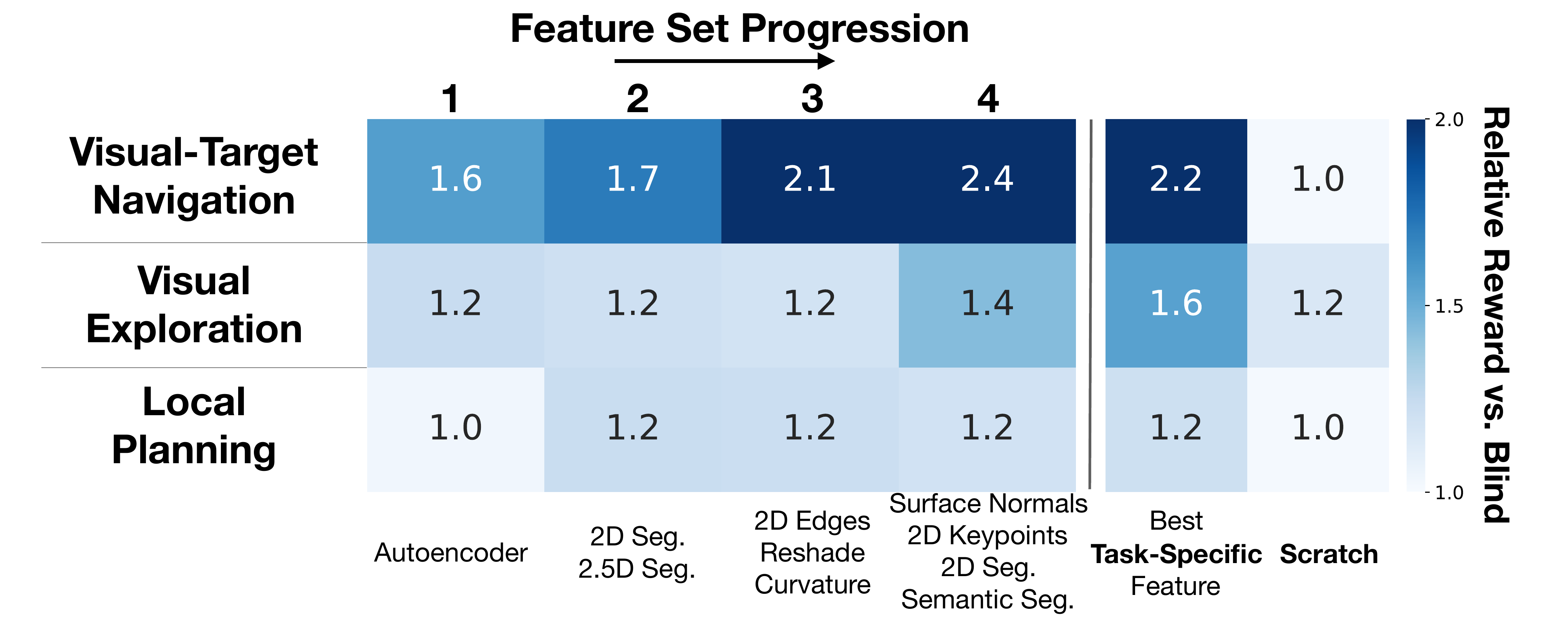}
        \vspace{-3.5mm}
        \caption{\footnotesize{\textbf{Evaluation of max-coverage feature sets.} Each cell denotes the reward (relative to \emph{blind}). The left 4 columns show agents trained with progressively larger max-coverage feature sets. The 4-feature set performs about as well as the best task-specific single feature---much better than the alternative approaches.}}
        \label{fig:perception_module_results}
    \vspace{-3mm}
\end{wrapfigure}

Since no single-objective representation could support all downstream tasks, we examined whether a \textbf{set} of single-objective representations could do better. We evaluated the Max-Coverage feature set for this purpose. Though it was derived in a way agnostic to the choice of task, we found that it performed nearly as well as the best feature for each task--\emph{as if we had known which feature to select}  (Fig.~\ref{fig:perception_module_results}). Training agents for Local Planning in Habitat and using M-C features sets with 2, 3, or 4 features yielded SPLs of 0.730, 0.733, 0.749, respectively, while the best mid-level feature yielded 0.76 SPL. Using a set of features was crucial, as choosing the best task-agnostic \emph{single} feature (\emph{autoencoder}) dropped SPL to 0.58. \emph{Scratch} and the SotA representation learning methods all  scored under 0.57 SPL. We found similar behavior on when training agents for other tasks in the Gibson environment~\ref{fig:perception_module_results}. The M-C feature set (with multiple features) also conferred the desirable emergent behaviors discussed in Sec.~\ref{section:generalization}, and agents using this set exhibited some of the lowest rates of collision, acceleration, and jerk. We include the full discussion and formulation of the M-C feature set and detailed experiments (e.g. M-C vs. alternative feature sets) in Appendix~\ref{apx:max_coverage_feature_set_analysis}.




\section{Conclusion}
\label{sec:conclusion}
\vspace{-2mm}   
In this paper we showed that one of the primary challenges with learning visuomotor policies is how to represent the visual input. We showed raw pixels are unprocessed, high dimensional, noisy, and difficult to work with. We presented an approach for representing pixels using \emph{mid-level visual representations} and demonstrated its utility in terms of improving final performance, boosting generalization, reducing sample complexity---significantly pushing the state-of-the-art. We also showed that the choice of representation generally depends on the final task. To this end we proposed a principled, task-robust method for computationally selecting a set of features, showing that the solver-selected sets outperformed state-of-the-art representations---simultaneously using an order of magnitude less data while achieving higher final performance.

\vspace{-0mm}
\textbf{Acknowledgements:} This material is based upon work supported by ONR MURI~(N00014-14-1-0671), Google Cloud, NSF (IIS-1763268), NVIDIA NGC beta, and TRI. Toyota Research Institute (``TRI'')  provided funds to assist the authors with their research but this article solely reflects the opinions and conclusions of its authors and not TRI or any other Toyota entity.


\footnotesize{
\bibliography{egbib}  
\nocite{kumar2018visual}
}


\newpage
\appendix

\title{Learning to Navigate Using Mid-Level Visual Priors\\ \vspace{3mm}\large{Appendix}}
\makesupmattitle

\large{\textbf{The following items are provided in the appendices:} }
\vspace{-1mm}
\normalsize

\startcontents[sections]
\printcontents[sections]{}{1}{}

\newpage


\renewcommand{\labelenumii}{\Roman{enumii}}

\newpage
\label{apx:first_page}
\section{Detailed Methodology}\label{apx:detailed_methodology}

\subsection{Baseline Description}\label{apx:baseline_description}
\label{sec:baselines}

In the main paper we described the only the most crucial baselines. We now provide descriptions for all baselines:

\begin{description}[leftmargin=2mm]
\small{

\item \textbf{Tabula Rasa (Scratch) Learning:}
The most common approach, \emph{tabula rasa} learning trains the agent from scratch. In this condition (sometimes called \emph{scratch}), the agent receives the raw RGB image as input and uses a randomly initialized AtariNet~\cite{mnih-dqn-2015} network (described in supplementary material). 

\vspace{-1mm}
\item \textbf{Blind Intelligent Actor:}
The \emph{blind} baseline is the same as \emph{tabula rasa} except that the visual input is a fixed image and does not depend on the state of the environment. The \emph{blind} agent is a particularly informative and crucial baseline since it indicates how much performance can be squeezed out of the nonvisual biases, correlations, and overall structure of the environment. For instance, in a narrow straight corridor which leads the agent to the target, there should be a small performance gap between \emph{sighted} and \emph{blind}. 

\vspace{-1mm}
\item \textbf{Random Nonlinear Projections:}
this is identical to using \emph{mid-level} features, except that the feature encoder is not pretrained but rather randomly initialized and then frozen. The  policy then learns on top of this fixed nonlinear projection.
This addresses the possibility that the ResNet architecture, not the offline perception task, is responsible the representations' success. 

\vspace{-1mm}
\item \textbf{Pixels as Features:}
this is identical to using \emph{mid-level} features, except that we downsample the input image to the same size as the features ($16\times 16$), apply a convolutional layer to match output sizes, and use it as the feature.
This addresses whether the feature readout network could be an improvement over AtariNet~\cite{mnih-dqn-2015}. 
}

\vspace{-1mm}
\item \textbf{Random Actions:} this uniformly randomly samples from the action
space. It calibrates how difficult the task is without any specialized method and determines how much can be obtained just from random chance. 

\vspace{-1mm}
\item \textbf{Non-Learning Methods:} For local planning, we compare against Simultaneous Localization and Mapping (SLAM) \cite{RGBDslamreview}, a popular approach that is highly effective when depth information is accurate. However, a depth sensor is not available and we estimate depth using the same network that our mid-level representation uses for distance.

\vspace{-1mm}
\item \textbf{State-of-the-Art Representation Learning:} We compare against several state-of-the-art representation-learning methods. They are not necessarily vision-centric, and they include dynamics-modeling ~\cite{Munk2016Forward, ShelhamerMAD16Inverse, Jordan1992ForwardMS}, curiosity~\cite{curiosity}, DARLA~\cite{higgins2017darla}, and ImageNet pretraining~\cite{alexnet}.

\end{description}

\subsection{Max-Coverage Feature Set: Formulation}\label{apx:max_coverage_feature_set_formulation}
As we showed in the main paper, no single representation could be universal, necessitating the use of a set of representations. Employing a larger set maximizes the chance of having the correct representation for the downstream task available and in the set. However, a compact set is desirable since agents using larger sets need more data to train (for the same reason that training from raw pixels requires many samples).
Therefore, we use a \textbf{Max-Coverage Feature Selector} that curates a compact subset of representations in order to ensure the \emph{ideal} representation (encoder choice) is never too far away from one in the set. 

\begin{figure}[h]
\centering
\vspace{-3mm}
\hspace{2mm}
\includegraphics[width=0.6\columnwidth]{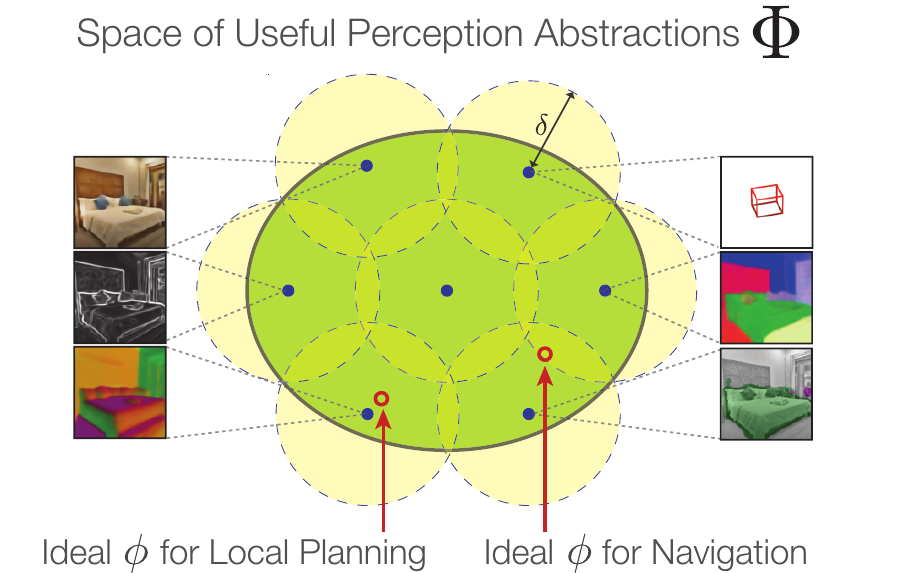}
\caption{\footnotesize{\textbf{Geometry of the feature set.} We select a covering set of features that minimizes the worst-case distance between the subset and the \emph{ideal} task feature. By \emph{Hypothesis III}, no single feature will suffice and a set is required. okay.}}\label{fig:perception_module}
\vspace{-1mm}
\end{figure}

The question now becomes how to find the best compact set, shown in Figure~\ref{fig:perception_module}. With a measure of distance between features, we can explicitly minimize the worst-case distance between the best feature and our selected subset (the \emph{perceptual risk} by finding a subset $X_{\delta} \subseteq \Phi = \{\phi_1, ..., \phi_m\}$ of size $|X_{\delta}| \leq k$ that is a $\delta$-cover of $\Phi$ with the smallest possible $\delta$. This is illustrated with a set of size 7 in Figure~\ref{fig:perception_module}.

The task taxonomy method~\cite{taskonomy2018} defines exactly such a distance: a measure between perceptual tasks. Moreover, this measure is predictive of (indeed, derived from) transfer performance. Using this distance, minimizing worst-case transfer (\emph{perceptual risk}) can be formulated as a sequence of Boolean Integer Programs (BIPs), parameterized by a boolean vector $x$ indicating which features should be included in the set.

This section describes the full sequence Boolean Integer Program that yields, as its solution, the Max-Coverage Min-Distance Feature Set. It accounts for interactions between feeatures. For example, what if the combination of two features is much stronger than either feature separately? This section details a variant of the main BIP that handles these interactions. It selects a set of transfers which \emph{may have one or more sources}.

The set of all transfers (edges), $E$, is indexed by $i$ and each edge has the form $(\{s^i_1, \dotso, s^i_{m_i}\}, t^i)$ (for $\{s^i_1, \dotso ,s^i_{m_i}\} \subset \mathcal{S}$ and $t^i \in \mathcal{T}$). We shall also index the target features $\mathcal{\Phi}$ by $j$ so that in this section, $i$ indexes edges and $j$ indexes target features. As in \cite{taskonomy2018}, we define operators returning target and sources of an edge:

\begin{align*}
\big(\{s^i_1, \dotso, s^i_{m_i}\}, t^i\big) \hspace{2mm} & \xmapsto[]{sources} \hspace{1mm} \{s^i_1, \dotso, s^i_{m_i}\}\\
\big(\{s^i_1, \dotso, s^i_{m_i}\}, t^i\big) \hspace{2mm} & \xmapsto[]{target} \hspace{4mm} t^i.
\end{align*}

We encode selecting a feature $t$ to include in the set as including the transfer $\big(\{t\}, t\big)$. 

The arguments of the problem are
\begin{enumerate}
    \item $\delta$, a given maximum covering distance
    \item $p_i$, a measure of performance on a target from each of its transfers (i.e. the affinities from~\cite{taskonomy2018})
    \item $x \in \mathbb{R}^{|E| + |\Phi|}$, a boolean variable indicating whether each transfer and each feature in our set
\end{enumerate}

The BIP is parameterized by a vector $x$ where each transfer and each feature in our set is represented by a binary variable; $x$ indicates which nodes are selected for the covering set and a satisfying minimum-distance transfer from each unselected feature to one in the set. The canonical form for a BIP is: 
\begin{align*}
\text{minimize: } & \mathbbm{1}^T x\,,\\
\text{subject to: } & Ax \preceq b \text{ and } x \in \{0, 1\}^{|E| + |\Phi|}\,.
\end{align*}

Each element $c_i$ for a transfer is the product of the importance of its target task and its transfer performance:
\begin{equation}
c_i := r_{\textit{target}(i)} \cdot p_i\,.
\end{equation}
Hence, the \emph{collective} performance on all targets is the summation of their individual AHP performance, $p_i$, weighted by the user specified importance, $r_i$.

Now we add three types of constraints via matrix $A$ to enforce each feasible solution of the BIP instance corresponds to a valid subgraph for our transfer learning problem: \emph{Constraint I:} no feature is too far away from the covering set, \emph{Constraint II:} if a transfer is included in the subgraph, all of its source nodes/tasks must be included too, \emph{Constraint III:}  each target task has exactly one transfer in. 

\emph{Constraint I: No feature is too far}. We ensure that this by disallowing edges whose ``distance'' is too large. In our case, we use affinities (so that higher values indicate a better transfer). Therefore, we filter out edges with small affinities by redefining $$E \triangleq \{~e \in E~|~\text{affinity}(e) \leq \delta~\}.$$

\emph{Constraint II: All necessary sources are present}.
For each row $a_i$ in $A$ we require $a_i \cdot x \leq b_i$, where 
\begin{align}
a_{i,k} &\defeq \begin{cases}
|\textit{sources}(i)| & \text{if } k = i\\
-1 & \text{if } (k - |E|) \in \textit{sources}(i)\\
0 & \text{otherwise}\\
\end{cases}\\
b_{i} &= 0.
\end{align}

\emph{Constraint II: One transfer per feature}.
Via two rows $a_{|E| + j}$ and $a_{|E| + j + 1}$, we enforce that target $j$ has exactly one transfer:
\begin{equation}
\begin{array}{l}
    a_{|E| + j, i} \hspace{10.2pt} \defeq \hspace{7.3pt} \mathbbm{1}_{\{\textit{target}(i)=j\}},\hspace{0.5cm} b_{|E| + j} \hspace{10.2pt} \defeq +1, \\ 
    a_{|E| + j+1, i} \defeq -\mathbbm{1}_{\{\textit{target}(i)=j\}},\hspace{0.5cm} b_{|E| + j+1} \defeq -1.
\end{array}
\end{equation}

Those elements of A not explicitly defined above are set to 0. The problem is now a valid BIP and can be optimally solved in a fraction of a second~\cite{gurobi}.

The above program finds a (not necessarily unique) minimum-size covering set with a covering distance at most $\delta$. Given that our feature set has only a finite number of distances, we can find the smallest $\delta$ by solving a sequence of these BIPs (e.g. with binary search). Since there are only $m^2$ distances, we can find the minimum $\delta$ with binary search, by solving $\mathcal{O}(log(m))$ BIPs. This takes under 5 seconds and the final boolean vector $x$ specifies the feature set of size $k$ that minimizes perceptual risk.

\subsection{Downstream Active Tasks}
\label{sec:supmat_tasks}

\begin{figure}[h]
\centering
\includegraphics[width=0.8\columnwidth]{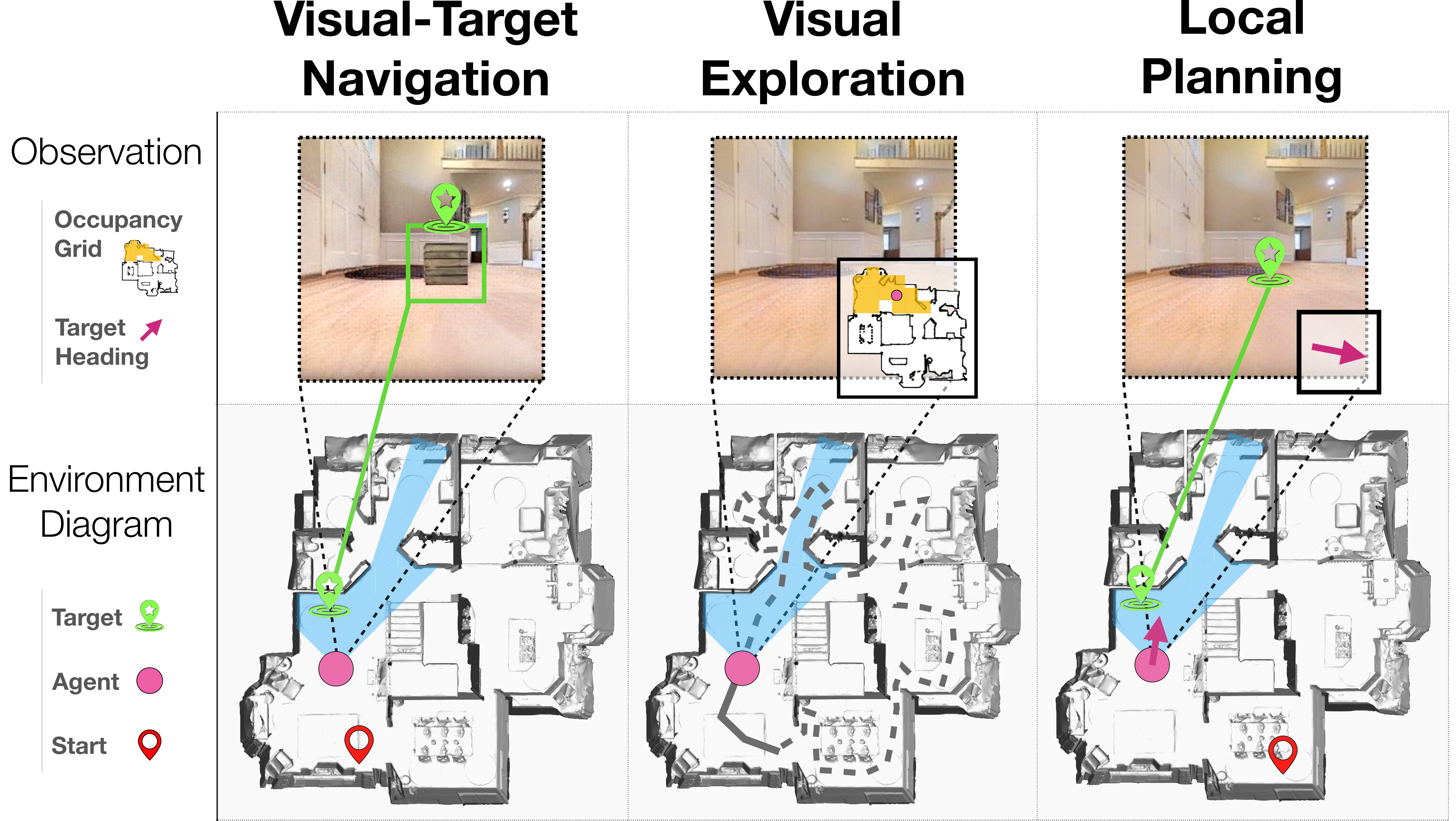}
\caption{\footnotesize{\textbf{Visual descriptions of the selected active tasks.} For each task, an example state of the environment is shown in the bottom row. The agent is shown as a pink dot, and its field of view is shown in light blue. Starting locaitons are shown with the red marker and a target, if it exists for the task, is shown with the green marker. The top row shows the corresponding sensory inputs, which always include some frame from the onboard RGB camera. Some tasks (local planning, visual exploration) include a additional nonvisual inputs, and these are shown in the bottom left corner of the RGB input (e.g. occupancy grid, target heading). Note that exploration uses only the revealed occupancy grid and no actual mesh boundaries.}} 
\label{fig:task_definition}
\end{figure}

This section contains detailed descriptions of our 3 locomotion-based active tasks. Visual depictions are shown in Fig.~\ref{fig:task_definition}, and (environment-specific numbers are provided in \href{http://perceptual.actor/#paper}{supplementary material}).

\subsubsection*{Local Planning:}
The agent must navigate to some target destination which is specified completely nonvisually (e.g. as a coordinate). This task is sometimes called ``pointnav'' or ``point navigation''.
    
    The reward function is dense with several terms: a single positive value when the goal is reached (which also terminates the episode), a small positive value per timestep for progress towards the goal (scaled change in distance to goal), a small negative value per timestep as a penalty for living and, for some environments, a larger negative value for obstacle collision. The reward function reflects desirable behavior of a \emph{local} planner, which should skillfully maneuver its environment while taking the most efficient path to the goal. We implement a sparsified variant which only contains the one-time bonus and the living penalty (see the main paper for experiments)
    
    At each timestep, the agent observed both the RGB camera images and a target vector $[r, \cos{\theta}, \sin{\theta}] \in \mathbb{R}^3$ where $(r, \theta)$ is the polar coordinates of the target in the agent coordinate system. In Habitat, we sometimes also provide a bitmap of past agent locations (to obviate the need for recurrence). The episode terminates either when the goal is reached or after a certain number of steps (usually 500). Sample frames are shown below and in the supplementary material. 
    
\subsubsection*{Visual Exploration:} The agent is equipped with a myopic laser scanner and must use it to explore as much of the space as possible in a limited amount of time (usually 1000 timesteps). The environment is partitioned into $1m \times 1m$ occupancy cells and the reward at every timestep is the number of occupancy cells newly uncovered by the laser scanner. 
    
    The cells that the agent sees are calculated via a ``myopic laser scanner'' in the following way: first forming a point cloud using a narrow strip of the depth image from the midpoint of the image to the bottom, and then projecting this point cloud onto the ground plane. To determine what cells are newly uncovered, we compare this against previously unlocked cells.
    
    At each timestep, the agent observed only the input frame from an onboard RGB camera and also the occupancy cells unlocked so far (provided as a bitmap image). The episode terminates automatically after a fixed number of agent steps.
    

\subsubsection*{Visual-Target Navigation:} In this task the agent must navigate to some object, specified visually. The agent must learn to both identify the target object as well as figure out how to navigate to it. The object (e.g. a wooden crate) remains fixed over training, but the agent and target locations are randomized.
  
    The reward is sparse, with a large one-time positive bonus for reaching the object and a otherwise small negative penalty of living. At each timestep, the agent received the RGB input frame from the onboard RGB camera and nothing else. The episode terminates when the target is reached or after a certain number of steps (usually 500). A sample frame is shown below, and more frames are shown in the supplementary material. 


\subsection{Dictionary of Mid-Level Vision Objectives }\label{sec:task_dict}
\begin{figure} [H]
  \centering
  \begin{minipage}{0.45\columnwidth}
  \centering
  \small
  
  \begin{tabular}{ |c| } 
    \hline
    \\
    \textbf{Mid-Level Vision Objectives ($\widetilde{\Phi}$)} \\[3ex]
    \hline
    \hline
    Autoencoding \\
    Classification, Semantic (1000-class) \\
    Classification, Scene \\
    Context Encoding (In-painting) \\
    Content Prediction (Jigsaw) \\
    Depth Estimation, Euclidean \\
    Edge Detection, 2D \\
    Edge Detection, 3D \\
    Keypoint Detection, 2D \\
    Keypoint Detection, 3D \\
    Reshading \\
    Room Layout Estimation \\
    Segmentation, Unsupervised 2D \\
    Segmentation, Unsupervised 2.5D \\
    Segmentation, Semantic \\
    Surface Normal Estimation \\
    Vanishing Point Estimation \\
    \hline 
    
  \end{tabular}
  \end{minipage}
  \begin{minipage}{0.45\columnwidth}
    \centering
    \begin{minipage}{0.72\columnwidth}
        \includegraphics[width=\textwidth]{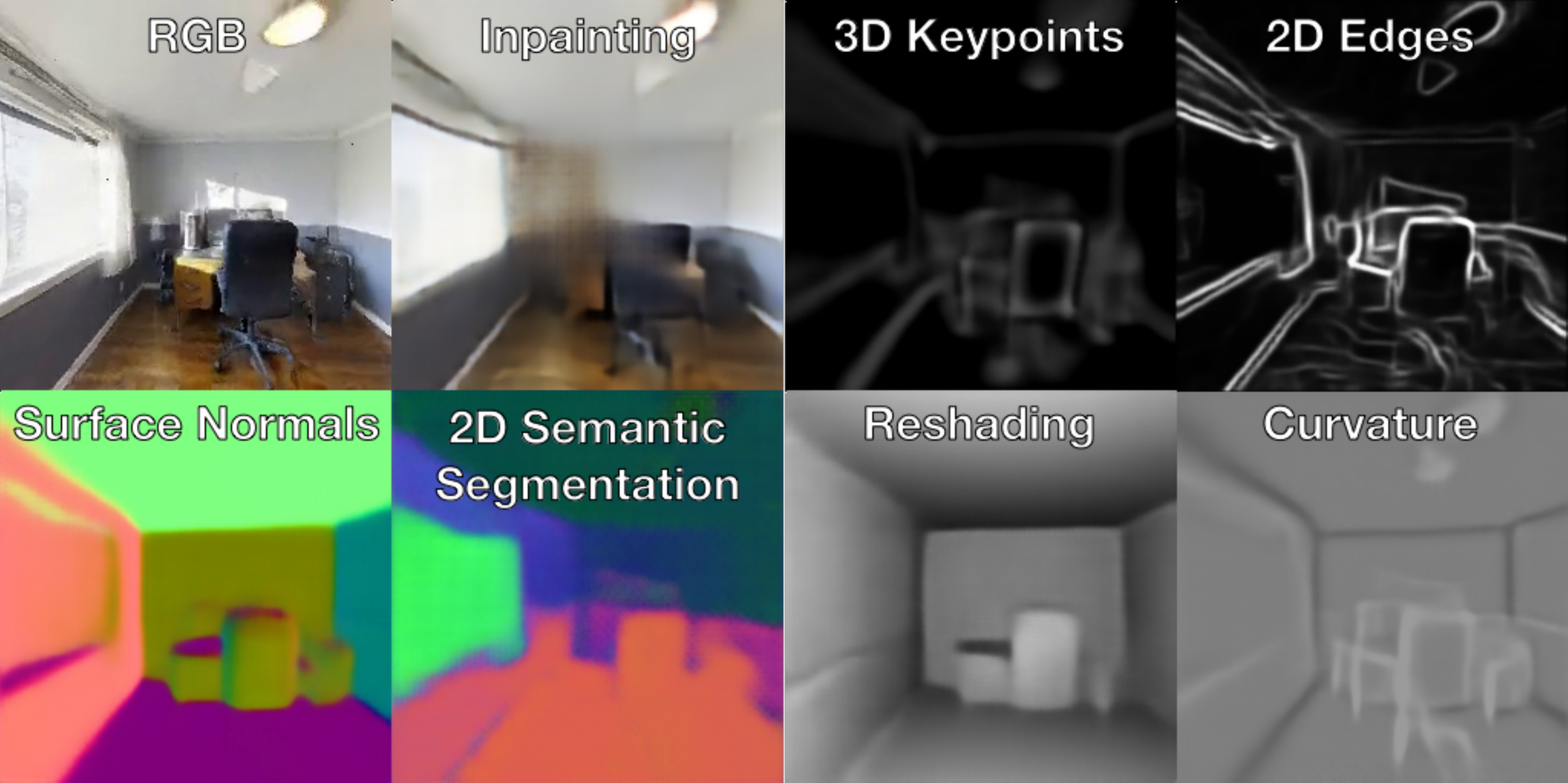}
    \end{minipage}
    \vspace{1mm}
    \begin{minipage}{0.72\columnwidth}
        \includegraphics[width=\textwidth]{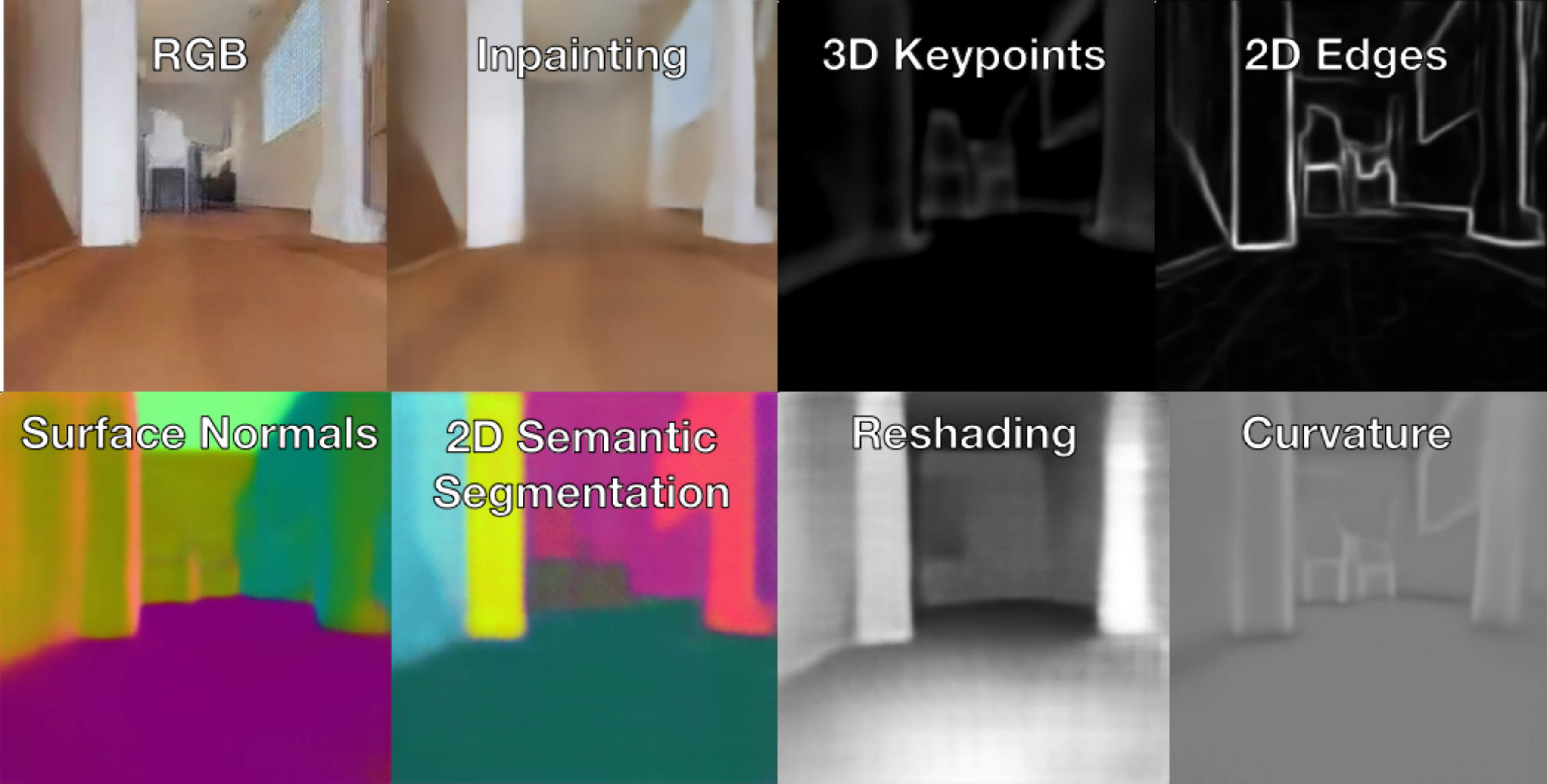}
    \end{minipage}
    \vspace{1mm}
    \begin{minipage}{0.72\columnwidth}
        \includegraphics[width=\textwidth]{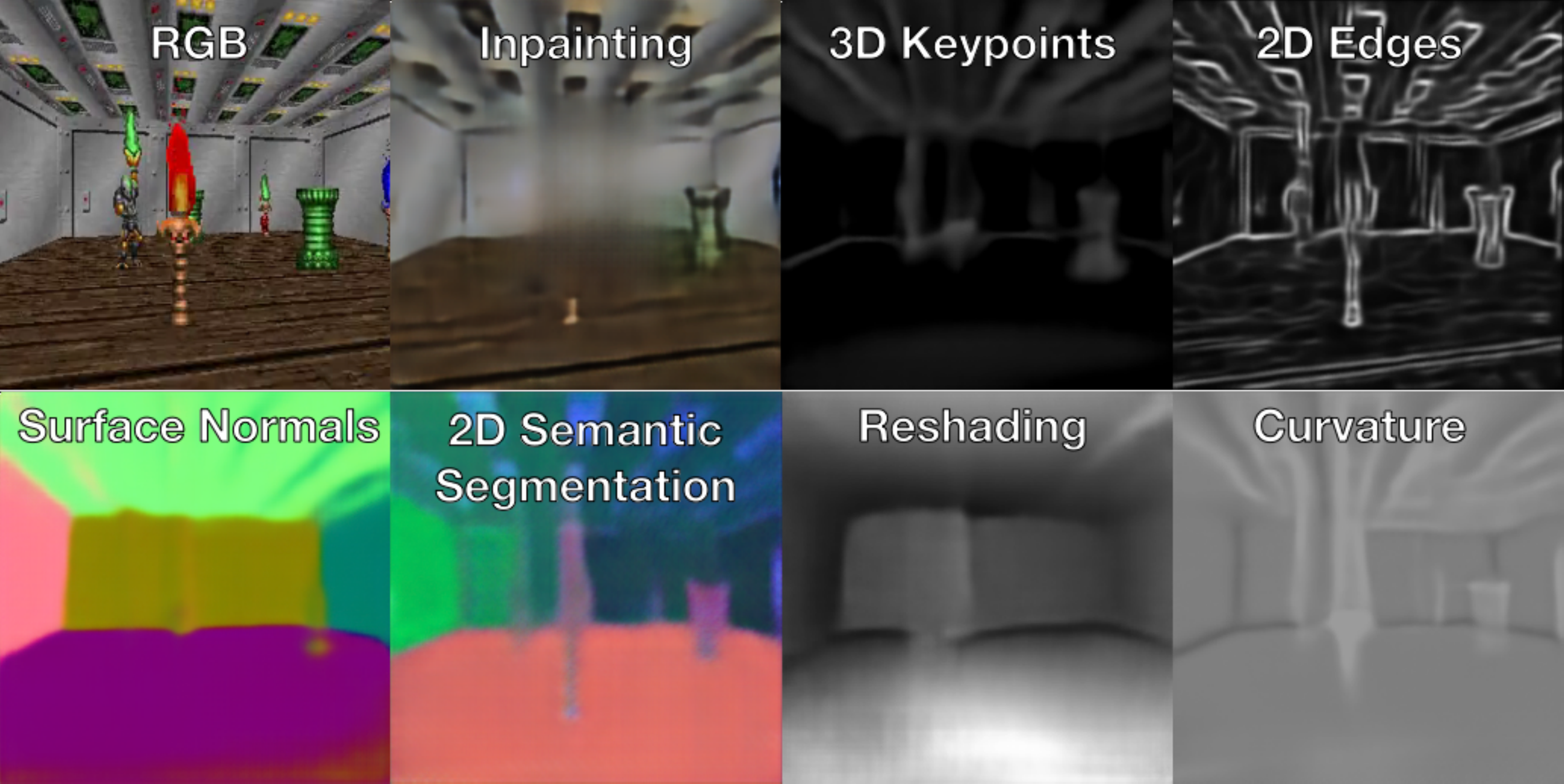}
    \end{minipage}
  \end{minipage}{}
  
  \vspace{0.5mm}
  \caption{\footnotesize{\textbf{Feature Bank with Sample Frames.} [Left] The mid-level features we used for all experiments. [Right] Frames and their respective mid-level features for Habitat, Gibson, and Doom for the top, middle, bottom frames respectively.}}
  \label{fig:supmat_taskbank}
\end{figure}

Figure~\ref{fig:supmat_taskbank} contains a complete list of our studied vision objectives. A detailed description of each vision task can be found the Taskonomy~\cite{taskonomy2018} supplementary material (Section 14). We visualize some tasks in Figure \ref{fig:supmat_taskbank} as well.

\subsection{Metrics} 
\label{sec:supmat_metrics}
\textbf{Reward Relative to Blind} RL results are typically communicated in terms of absolute reward. However, absolute reward values are uncalibrated and a high value for one task is not necessarily impressive in another. One way to calibrate rewards according to task difficulty is by comparing to a control that cannot access the state of the environment. Therefore, we propose the \textit{reward relative to blind}: \vspace{-2mm}

\begin{equation}
RR_{\text{blind}} = \frac{r_\text{treatment} - r_{\text{min}}}{r_\text{blind} - r_{\text{min}}}
\end{equation}
as a calibrated quantification. A \emph{blind} agent always achieves a relative reward of 1, while a score $>1$ indicates a relative improvement and score $<1$ indicates this agent performs worse than a \emph{blind} agent. We find this quantification particularly meaningful since we found agents trained from scratch often memorize the training buldings, performing no better than \emph{blind} in the test setting (see Section \ref{section:generalization} in main paper). For completeness, we provide the raw reward curves below in supplementary material. 

\textbf{Success Weighted by Path Length} For local planning, we also use the metric of Success weighted by (normalized inverse) Path Length (SPL) introduced by \cite{anderson2018evaluation}. The measure is defined as follows: 
\begin{quote}
    We conduct N test episodes. In each episode, the agent is tasked with navigating to a goal. Let $l_i$ be the shortest path distance from the agent's starting position to the goal in episode $i$ and let $p_i$  bet he length of the path actually taken by the agent. Let $S_i$ be a binary indicator of success in episode $i$. We define a summary measure of the agent's navigation performance across the test set as follows: 
    $$\frac{1}{N}\sum_{i=1}^{N} S_i \frac{l_i}{max(p_i, l_i)}.$$
\end{quote}{}

\subsection{PPO with Experience Replay}
\label{apx:ppo}
In this section we describe our off-policy PPO variant which decorrelates the samples within each batch and provides more stable, sample-efficient learning. 

\subsubsection*{Off-Policy Policy Gradient}

In the most general policy gradient setup, we attempt to maximize the objective function:

\begin{equation}
    \mathbb{E}_{\tau \sim \pi_{\theta}(\tau)} \Big[ \log \pi_{\theta}(a_t | s_t) \hat{A}_t \Big]
\end{equation}

where $\hat{A}_t$ is the advantage function at timestep $t$ (some sufficient statistic for the value of a policy at timestep $t$, in our experiments we choose the generalized advantage estimator \cite{gae}) and $\tau$ is a trajectory drawn from the current policy. The right side of the equation is our estimate of the objective using data sampled from the environment under the policy. Using importance sampling, we can approximate this objective by sampling from a different distribution over trajectories, and instead optimize:

\begin{equation}
    \mathbb{E}_{\tau \sim \pi_{\theta_{\text{old}}}(\tau)} \Big[ \frac{\pi_{\theta}(a_t | s_t)}{\pi_{\theta_{\text{old}}}(a_t | s_t)} \hat{A}_t \Big]
\end{equation}

\subsubsection*{Off-Policy PPO}

In practice, if we were to directly optimize the objective in equation (2), this would lead to dramatically unstable gradient updates due to both the high variance of both the importance sampling ratio and $\hat{A}_t$ (in our actor-critic framework, we are learning $\hat{A}$ as the critic as training progresses, so in the beginning of training $\hat{A}$ is especially unstable).

Proximal policy optimization decreases variance by clipping the policy ratio, minimizing the surrogate objective:

\begin{equation}
   \mathbb{E}_{\tau \sim \pi_{\theta_{\text{old}}}(\tau)} \min\Big[ r_t(\theta) \hat{A}_t, \text{clip}(1 - \epsilon, 1+\epsilon, r_t(\theta)) \hat{A}_t \Big]
\end{equation}

\begin{equation}
r_t(\theta) = \frac{\pi_{\theta}(a_t | s_t)}{\pi_{\theta_{\text{old}}} (a_t | s_t)}
\end{equation}

 This objective acts as a first-order trust region, preventing large policy updates that drastically change the policy. For detailed theoretical justification of this objective, see the PPO paper \cite{PPO}. In the original PPO paper, $\pi_{\theta_\text{old}}$ is only the distribution of on-policy trajectories from parallel workers. Since we are constrained to only one worker, sampling from on-policy trajectories only would lead to batches with highly correlated samples, leading to overfitting. Instead, we maintain a replay buffer, and draw multiple trajectories from the replay buffer at every update, treating the policy at the time when the trajectories were executed as $\pi_{\theta_\text{old}}$. To calculate the advantage, we reevaluate the advantage function during the update using the current critic on the states on the sampled trajectories from the replay buffer. The number of on-policy and off-policy trajectories that we sample under this formulation is a hyperparameter which we tune and report.

\newpage
\section{Additional Experiments and Analysis}\label{apx:additional_analysis}
We include more results on the desired properties of midlevel representation, namely better performance and stronger generalization. We look at experiments across multiple environments and multiple downstream tasks. 

\subsection{Performance} 

\begin{figure}[h]
    \centering
    \includegraphics[width=0.7\textwidth]{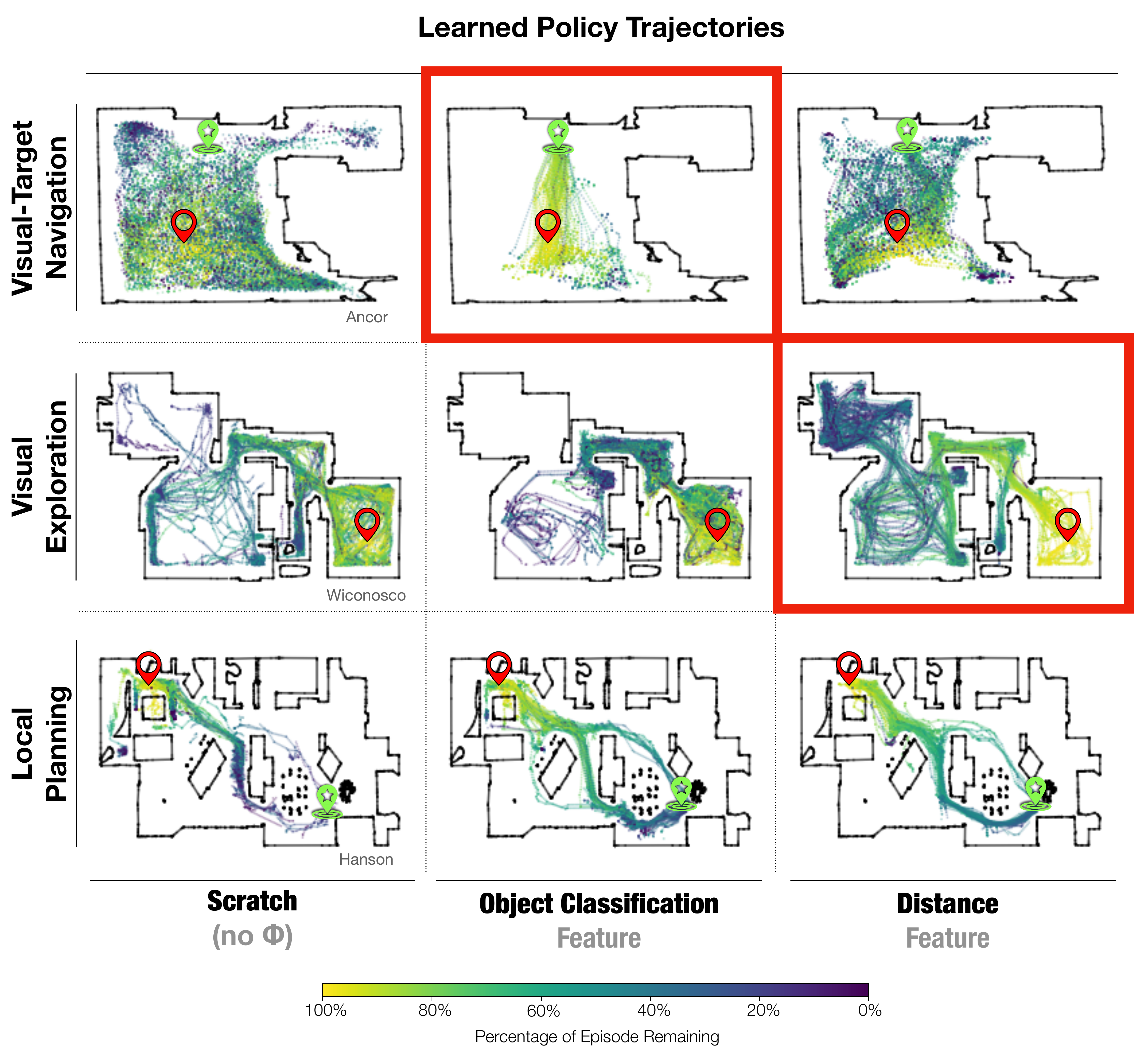}
    \label{fig:supmat_visual_trajectories}
    \caption{\footnotesize{\textbf{Visualizations of the agents' trajectory.} While all approaches have reasonable trajectories for local planning, only certain features have desired trajectories for visual-target navigation and visual exploration.}}
\end{figure}

Are the differences in the reward values meaningful? Do they led to different behaviors? In Figure \ref{fig:supmat_visual_trajectories}, we superimpose 100 evaluation trajectories in the three Gibson tasks of scratch and two different features, namely object classification and distance estimation. The scratch policy (left) completely fails to perform visual exploration or visual-target navigation with desirable trajectories,  inefficiently wandering around the test space. The policy trained with object classification (middle) recognizes and converges on the navigation target (boxed), but fails to cover the entire space in exploration. This is perhaps due to the fact that semantic information is not as useful for exploration.  Distance estimation features (right) help the agent cover nearly the entire space in exploration (boxed), but fail in navigation unless the agent is nearly on top of the target. This is perhaps due to the fact that geometric tasks fail to understand the semantics of the target.


\begin{figure}[ht]
    \centering
    \includegraphics[width=1.0\textwidth]{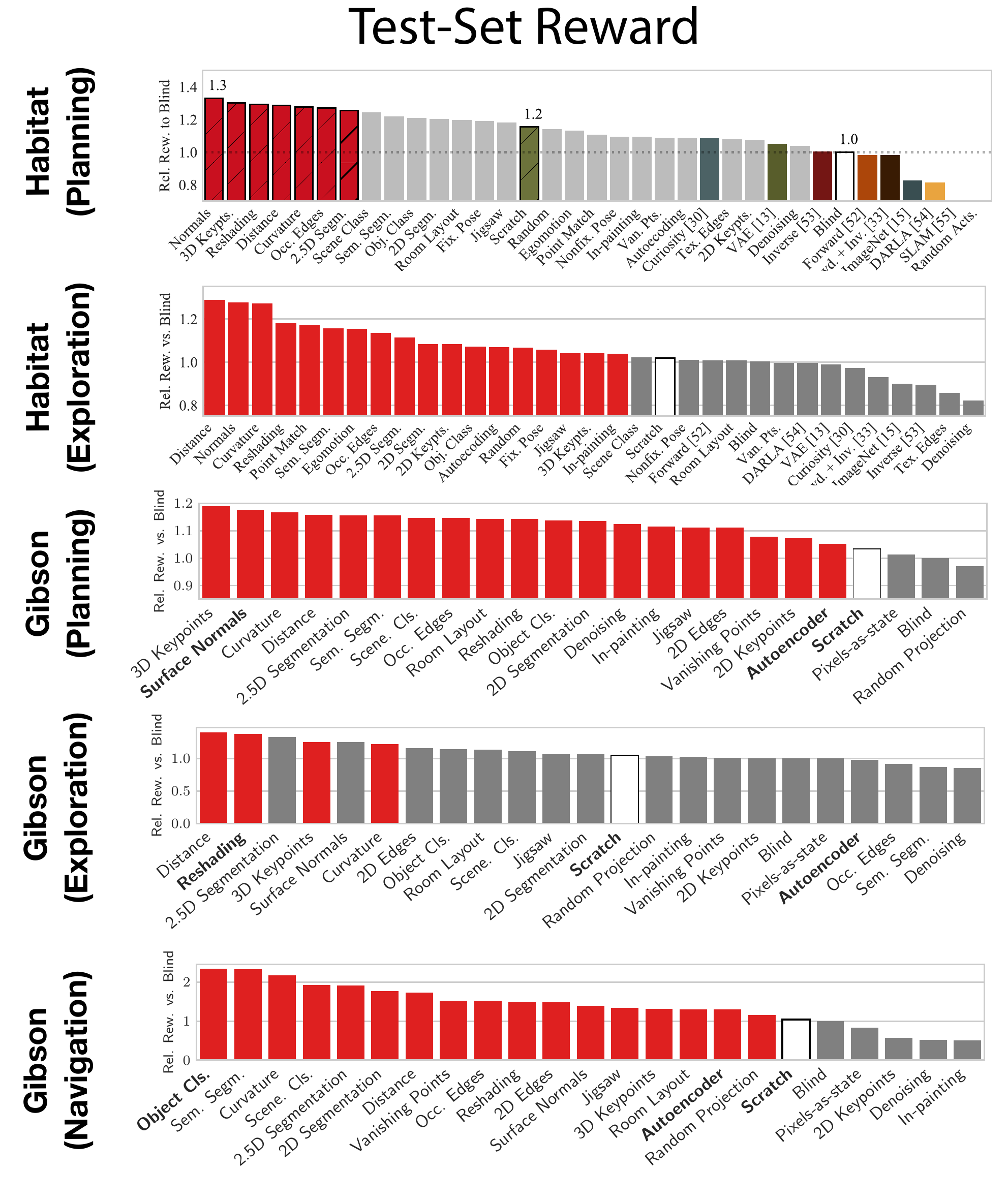}
    \label{fig:supmat_test_set_all_reward}
    \caption{\footnotesize{\textbf{Performance.} The reward relative to blind for all methods. Agents significantly better than \emph{scratch} are shown in \textcolor{red}{red}.}}
\end{figure}


Figure \ref{fig:supmat_test_set_all_reward} shows features that achieve a higher reward than scratch on a held-out set of validation buildings. Those features that do so with high confidence are highlighted in red
For each task, some features outperform \emph{tabula rasa} learning. We measure significance using the nonparametric Mann-Whitney U test, correcting for multiple comparisons by controlling the False Discovery Rate ($Q=20\%$) with Benjamini-Hochberg~\cite{Benjamini_hochberg}. These significance tests reveal that the probability of so many results being due to noise is $< 0.002$ per task ($< 10^{-6}$ after adding the seeds from the analysis in Section~\ref{apx:supmat_rank_reversal}).

\newpage
\subsection{Generalization}\label{apx:additional_generalization}
We investigate the generalization further in Gibson and Doom and show that our method generalizes better to new domains.


\subsubsection{Generalization to Gibson Test Buildings}\label{apx:additional_generalization_scratch_vs_blind}
\begin{figure}[h]
    \centering
    \includegraphics[width=\textwidth]{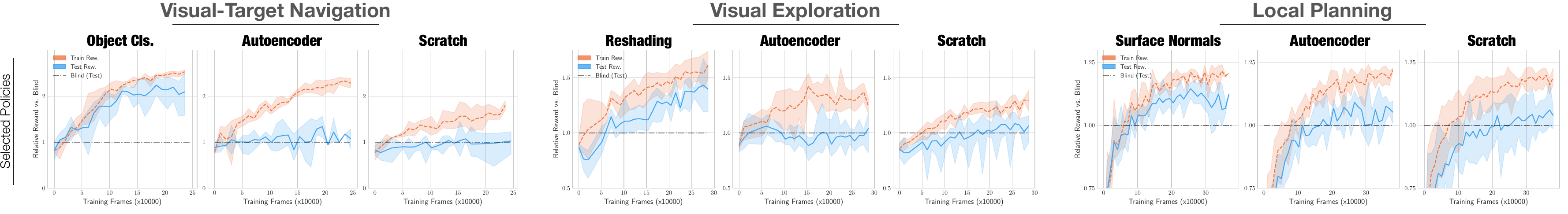}
    \vspace{0mm}
    \caption{\footnotesize{\textbf{Mid-level feature generalization.} The plots above show training and test performance of \textit{scratch} vs. some selected features throughout training. For all tasks there is a significant gap between train/test performance for \textit{scratch}, and a much smaller one for the best feature. This underscores the importance of separating the train and test environment in RL.}}
    \label{fig:supmat_gibson_generalization}
\end{figure}

We find that for each of our tasks, several feature-based agent not only achieved higher final test performance than policies trained \emph{tabula rasa} but also exhibited a smaller gap between training and testing performance. 
All agents exhibited some gap between training and test performance, but agents trained from scratch seem to overfit completely---rarely doing better than blind agents in the test environment. The plots in Figure~\ref{fig:supmat_gibson_generalization} show representative examples of this disparity over the course of training. Similarly, some common features like \textit{Autoencoders} and \emph{VAEs} have strong training curves that belie exceptionally weak test-time performance.

\subsubsection{Generalization to Texture Variation in Doom}
\begin{wrapfigure}{r}{0.4\textwidth}
    \vspace{-10mm}
    \centering
    \includegraphics[width=0.4\textwidth]{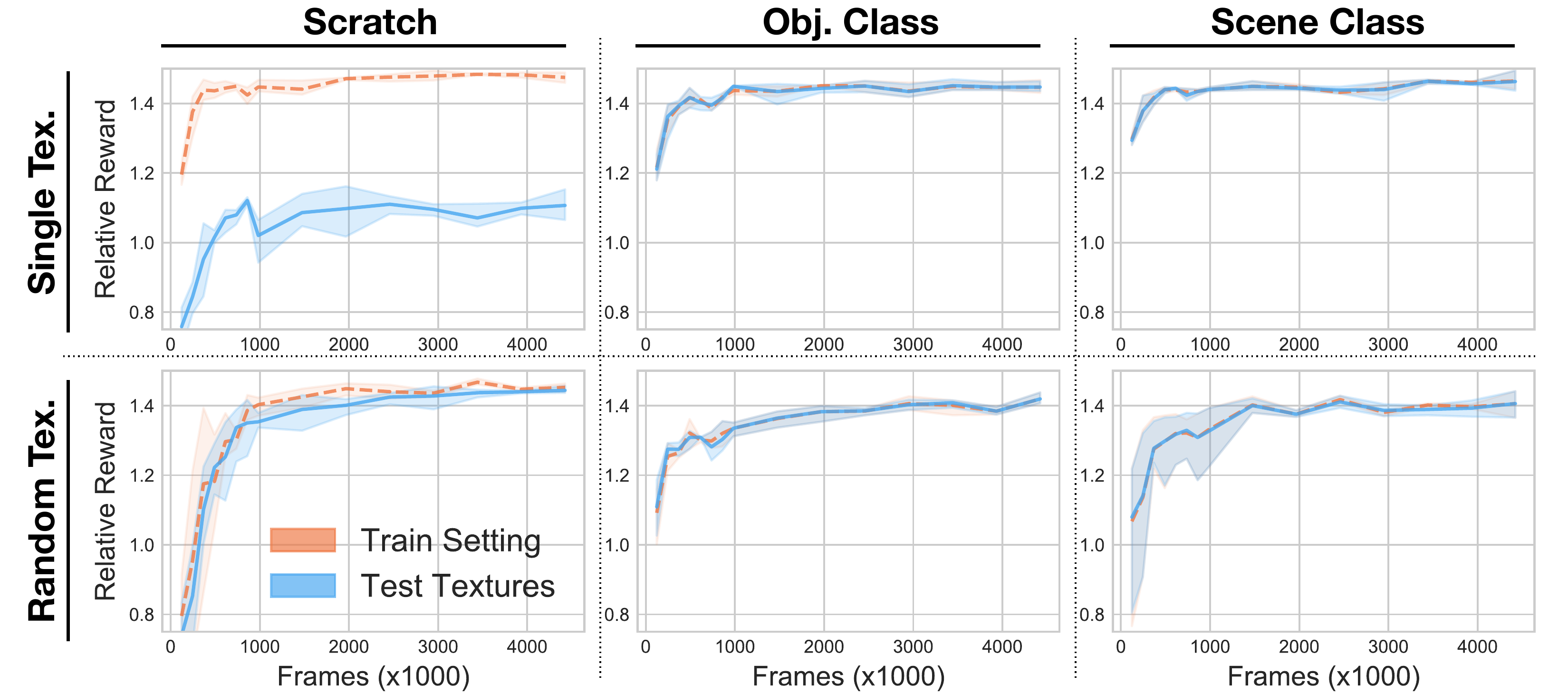}
    \caption{\footnotesize{\textbf{Generalization in Doom Textures} In ViZDoom, feature-based agents (right two columns) generalize to new textures even when not exposed to texture variation in training (top row), while agents trained from scratch suffer a significant drop in performance (left, top).}}
    \label{fig:supmat_doom_generalization}
    \vspace{-11mm}
\end{wrapfigure}{}
We also found that mid-level features were more robust than learning from scratch to changes in texture. While \emph{scratch} achieves the highest final performance when the agent learns in a video game environment where there are many train textures that emulate the test textures, \emph{scratch} fails to generalize when there is little or no variation in texture during training. On the other hand, feature-based agents were able to generalize even without texture randomization, as shown in Figure \ref{fig:supmat_doom_generalization}. 





\subsection{Rank Reversal}
\label{apx:supmat_rank_reversal}
In the main paper, we showed that geometric features were superior to semantic ones on exploration, but the opposite was true for visual navigation. We refer to this phenomenon as ``rank reversal'' - given any downstream task, the ranking of a feature is reversed on a different enough task. 
Here, we provide additional evidence of the ubiquity of rank reversal and offer evidence that there is no ``universal'' feature which maximizes performance for all active tasks. We start by looking at the rankings of features across tasks and follow with rigorous significance testing.

\subsubsection{Feature Rankings in different tasks}
\begin{wrapfigure}{r}{0.4\textwidth}
    \vspace{-13mm}
    \centering
    \includegraphics[width=0.4\textwidth]{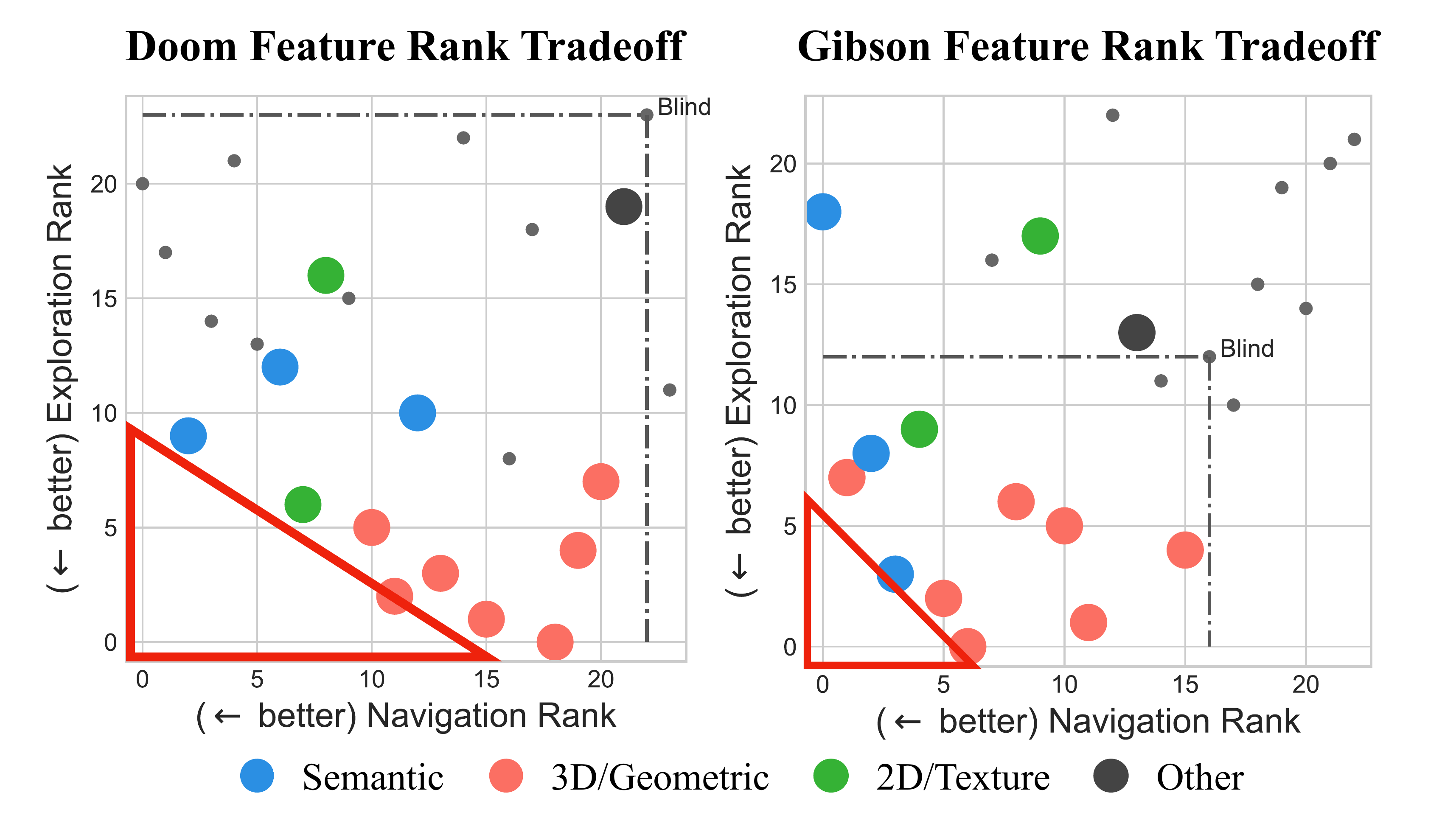}
    \caption{\footnotesize{\textbf{Feature Ranks on Exploration and Navigation} Scatterplots of the rank of feature performance on navigation (x-axis) and exploration (y-axis) in both Gibson (right) and Doom (left). The absence of any feature in the bottom left corners implies that there is no single universal feature. }}  
    \vspace{-5mm}
    \label{fig:submat_features_rank_tradeoff}
\end{wrapfigure}

In Figure \ref{fig:submat_features_rank_tradeoff}, we plot the rankings of all the mid-level features at exploration and navigation. The lack of any feature existing in the bottom left corner of the plots shows that there is no feature which ranks the very highly for both tasks. The performance of the mid-level visual features is highly dependent on the properties of the downstream task. These results, reproduced both in Gibson and in Doom, substantiates the idea that no mid-level feature will maximize reward a priori and thus there is no universal feature. We verify this observation using significance testing in the following section.

\subsubsection{Analysis of Geometric and Semantic Features across tasks}

\begin{figure}[H]
  \centering
  \begin{minipage}{0.49\columnwidth}
    \footnotesize
    \centering
    \begin{tabular}{ |c|c|c|c| } 
        \hline
        \textbf{Feature} & \textbf{Rew.} & \textbf{p-val.} &  $\mathbf{i/m}$ \\
        \hline
        \hline
        Sem. Segm.           &  6.553  &  0.0000  &  0.04 \\
        Scene Cls.           &  5.969  &  0.0001  &  0.08 \\
        Obj. Cls.            &  4.212  &  0.0004  &  0.12 \\
        Reshading            &  0.525  &  0.7824  &  0.16 \\
        3D Keypts.           &  -0.196  &  0.9460  &  0.20 \\
        \hline
        Distance             &  1.015  &  -  &  - \\
        \hline
    \end{tabular}
  \end{minipage}
  \begin{minipage}{0.49\columnwidth}
    \footnotesize
    \centering
    \begin{tabular}{ |c|c|c|c| } 
        \hline
        \textbf{Feature} & \textbf{Rew.} & \textbf{p-val.} &  $\mathbf{i/m}$ \\
        \hline
        \hline
        Distance             &  5.265  &  0.001  &  0.04 \\
        3D Keypts.           &  5.269  &  0.002  &  0.08 \\
        Reshading            &  5.072  &  0.011  &  0.12 \\
        Sem. Segm.           &  4.132  &  0.502  &  0.16 \\
        Scene Cls.           &  3.996  &  0.702  &  0.20 \\
        \hline
        Obj. Cls.            &  4.151  &  -  &  - \\
        \hline
    \end{tabular}
  \end{minipage}
  \vspace{0.5mm}
  \caption{\footnotesize{\textbf{Rank-reversal (Gibson).} [Left] We test whether features are significantly better than distance estimation (the best feature for exploration) in navigation. [Right] We test whether features are significantly better than object classification (the best feature for navigation) in exploration. While within families (semantic, geometric), the differences are not significant, across families, the differences are significant.}}
  \label{fig:supmat_reversal_sig_test}
\end{figure}


We compare the top performing exploration feature (distance) to other features in navigation and visa-versa (top performing navigation feature to other features in exploration).
For each feature and for each task, we train 10 agents from 10 random seeds and evaluate them at the end of training (420 updates for navigation, 480 updates for exploration). We evaluate the performance of each agent/seed combination by running the agent in the test environment for 100 random episodes. We then use a cluster-effect-adjusted Wilcoxon rank-sum test to test whether there is a significant difference in the reward-per-episode between features and show the results in Figure \ref{fig:supmat_reversal_sig_test}. 

\begin{wrapfigure}{l}{0.46\textwidth}
    \definecolor{navigation}{rgb}{0.19215686, 0.56078431, 0.89019608}
    \definecolor{exploration}{rgb}{0.98039216, 0.43529412, 0.38823529}
    \centering
    \includegraphics[width=0.46\textwidth]{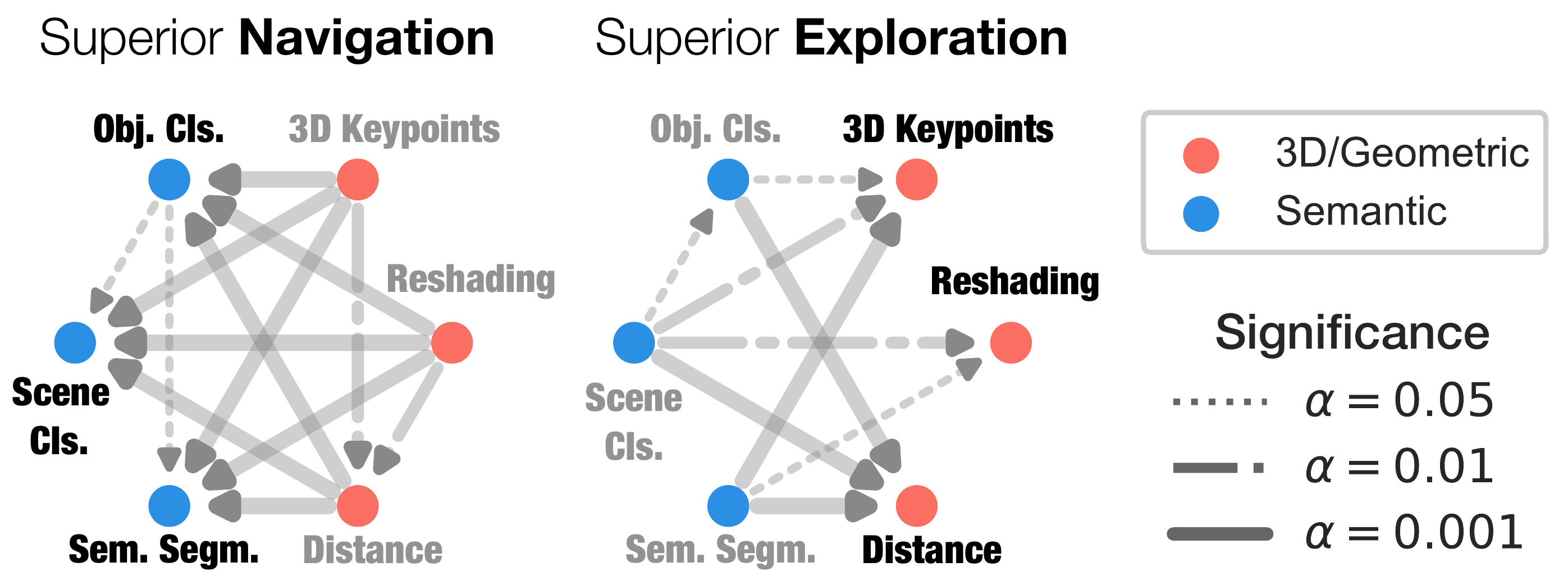}
    \caption{\footnotesize{\textbf{Rank Reversal Significance Graphs} Results from pairwise tests evaluating 3 \textcolor{navigation}{semantic} and 3 \textcolor{exploration}{geometric} features on each task. For each task graph, arrows point towards the better-performing feature. Lack of an arrow indicates the performance difference was not statistically significant. Heavier arrowheads denote more significant results (lower $\alpha$-level). The essentially complete bipartite structure in the graphs shows that navigation is characteristically semantic while exploration is geometric.}} 
    \vspace{-7mm}
    \label{fig:supmat_rank_reversal}
\end{wrapfigure}


We find that within each family of tasks (e.g. semantic, geometric), there is no significant difference. However, a family performing well on one active task always has a significantly lower reward-per-episode compared to the other family on the other active task. For example, distance is statistically worse than any semantic task at navigation, cementing the rank reversal hypothesis. 

In Figure \ref{fig:supmat_rank_reversal}, we perform pairwise significance tests between 3 well-performing mid-level features from each family. We find that the statistically different pairs come from different families. Additionally, we observe that while for navigation, arrows point heavily from geometric towards semantic tasks indicating significantly higher reward in the latter, the reverse is true for exploration (arrows point from semantic towards geometric tasks). Thus, there is no single task (or even family of tasks!) that consistently obtains significantly higher reward. We show results from Gibson but we came to the same conclusion in Doom. 



\subsection{Analysis of Max-Coverage Feature Set}\label{apx:max_coverage_feature_set_analysis}

This section contains additional analysis of the Max-Coverage Feature Set.

\begin{figure}[h]
\centering
\includegraphics[width=0.8\columnwidth]{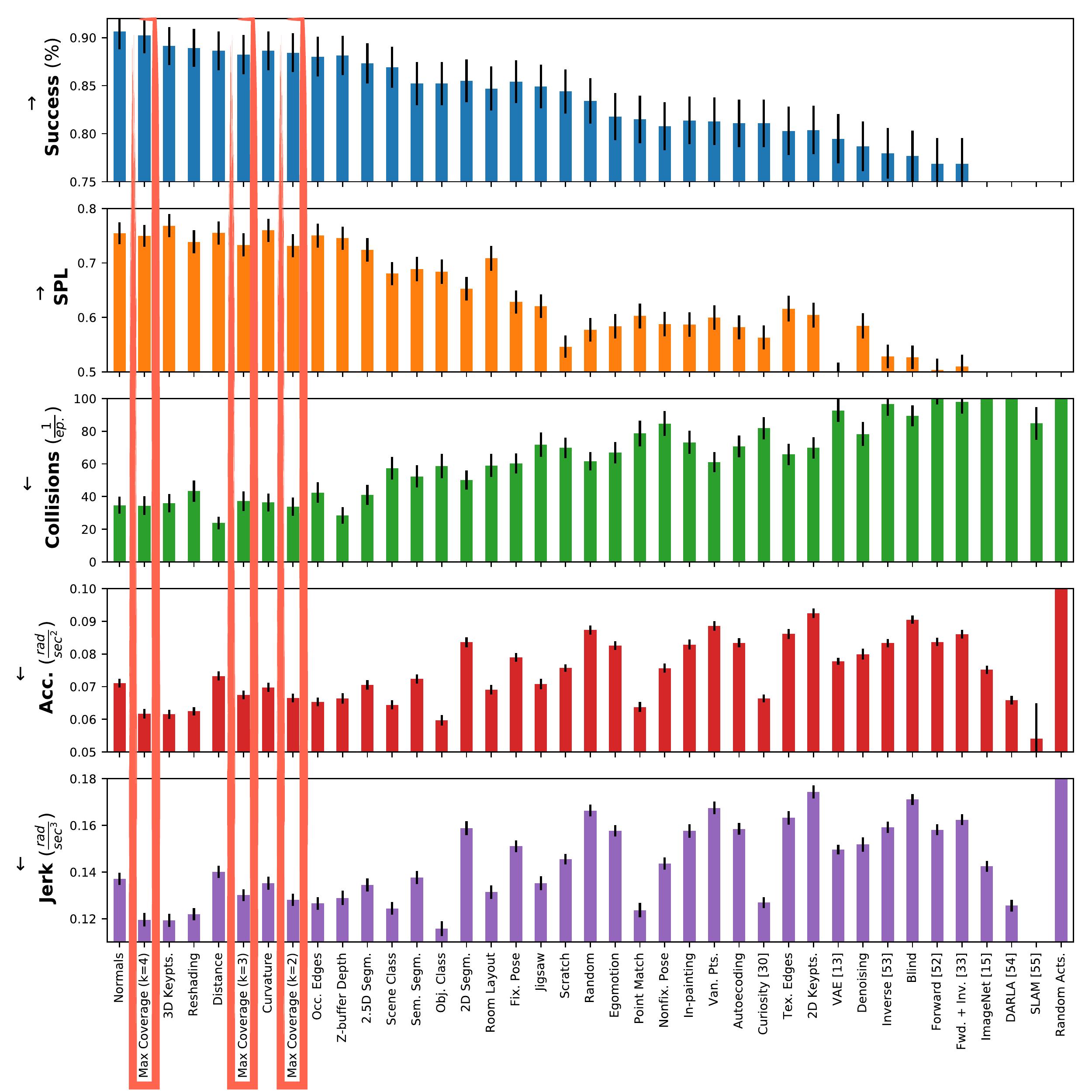}
\caption{\footnotesize{\textbf{Max-coverage feature sets exhibit strong performance and desirable behavior on Local Planning in Habitat.} The shows performance on the Habitat/Local Planning test set along a variety of dimensions. Features are ordered according to \emph{test-set reward}. Max-coverage policies exhibit a strong combination of desirable properties, suggesting that they confer the benefits of mid-level vision.}}
\label{fig:perception_module_analysis}
\end{figure}

\textbf{Max-Coverage Feature Set vs. Other Methods} Figure \ref{fig:perception_module_analysis} shows the findings (Success, SPL, Collisions, Acceleration, Jerk) for max-coverage feature set. The max-coverage feature set has similar performance with the best performing features while being task agnostic. 

\begin{table}[h]
\centering
\begin{tabular}{lllllll}
\hline
\multirow{2}{*}{Features} & \multicolumn{2}{l}{Navigation} & \multicolumn{2}{l}{Exploration} & \multicolumn{2}{l}{Planning} \\ \cline{2-7} 
                          & \textit{Ours}      & Rand.     & \textit{Ours}   & Rand.         & \textit{Ours}     & Rand.    \\ \hline
2                         & \textbf{1.7}       & 1.5       & 1.2             & \textbf{1.4}  & 1.2               & 1.2      \\ \hline
3                         & \textbf{2.1}       & 1.8       & 1.2             & \textbf{1.3}  & 1.2               & 1.2      \\ \hline
4                         & \textbf{2.4}       & 1.9       & \textbf{1.4}    & 1.3           & 1.2               & 1.2      \\ \hline
\end{tabular}
\caption{\footnotesize{\textbf{Max Coverage Feature Set outperforms random feature set (Gibson).} We compared the Max-Coverage feature set to random feature sets, and the M-C feature set performs better  than random feature sets. Each cell shows reward relative to blind.}}
\label{fig:random_feature_sets}
\end{table}

\textbf{Max-Coverage Representation Set vs. Random Feature Set:}How useful is the feature set proposed by the perception module? Will any feature set work? We randomly uniformly selected five feature sets and evaluated their performance on our active tasks.  Table~\ref{fig:random_feature_sets} shows that the solver-suggested feature set performs much better than the randomly selected sets on navigation, and comparably on the other two tasks. We hypothesize that the high performance of random features on exploration is due to a larger number of geometric-based tasks in our task dictionary, which tend to excel at the exploration task, while the relatively worse performance on navigation is due to the small number of semantic features in our dictionary. Our perception module ensures coverage of both kinds of features, which leads to good performance on both navigation and exploration.

It is notable that both our perception module and random sets of features outperform \emph{tabula rasa} learning (and our other baselines) by a wide margin. 

\newpage

\stopcontents[sections]
\end{document}